\documentclass[10pt,twocolumn,letterpaper]{article}

\usepackage[pagenumbers]{iccv} %
\newif\ifarxiv
\arxivtrue %

\usepackage{cuted} %
\usepackage{graphicx}
\usepackage{bm}

\renewcommand{\vec}[1]{\bm{#1}}
\newcommand{\mat}[1]{\mathbf{#1}}
\newcommand{\mesh}[1]{\mathcal{#1}}

\newcommand{\R}[1]{{%
    \textbf{%
        \ifstrequal{#1}{1}{\textcolor{teal}{R#1}}{%
        \ifstrequal{#1}{2}{\textcolor{blue}{R#1}}{%
        \ifstrequal{#1}{3}{\textcolor{magenta}{R#1}}{%
        \ifstrequal{#1}{4}{\textcolor{red}{R#1}}{%
                           \textcolor{cyan}{R#1}%
        }}}}%
    }%
}}

\definecolor{iccvblue}{rgb}{0.21,0.49,0.74}
\usepackage[pagebackref,breaklinks,colorlinks,allcolors=iccvblue]{hyperref}
\usepackage[capitalize]{cleveref}

\def\paperID{3337} %
\def\confName{ICCV}
\def\confYear{2025}

\title{HairCUP: Hair Compositional Universal Prior for 3D Gaussian Avatars}

\def\authorBlock{
    Byungjun Kim\textsuperscript{1,2}\thanks{Work done during the internship at Codec Avatars Lab, Meta} \qquad
    Shunsuke Saito\textsuperscript{2} \qquad
    Giljoo Nam\textsuperscript{2} \qquad
    Tomas Simon\textsuperscript{2} \\
    Jason Saragih\textsuperscript{2} \qquad
    Hanbyul Joo\textsuperscript{1}\textsuperscript{$\dag$} \qquad
    Junxuan Li\textsuperscript{2}\textsuperscript{$\dag$} \\[0.5em]
    \textsuperscript{1}Seoul National University \qquad
    \textsuperscript{2}Codec Avatars Lab, Meta \\
    {\tt\small \href{https://bjkim95.github.io/haircup/}{\color{magenta}{https://bjkim95.github.io/haircup/}}}
    \vspace{-5mm}
}

\author{\authorBlock}

\begin{document}
\maketitle

\renewcommand{\thefootnote}{\fnsymbol{footnote}}
\footnotetext[2]{Co-corresponding authors}

\begin{strip}
    \centering
    \vspace{-17mm}
    \begin{minipage}{\textwidth}
        \centering
        \includegraphics[width=0.95\textwidth,trim={0 0 0 0},clip]{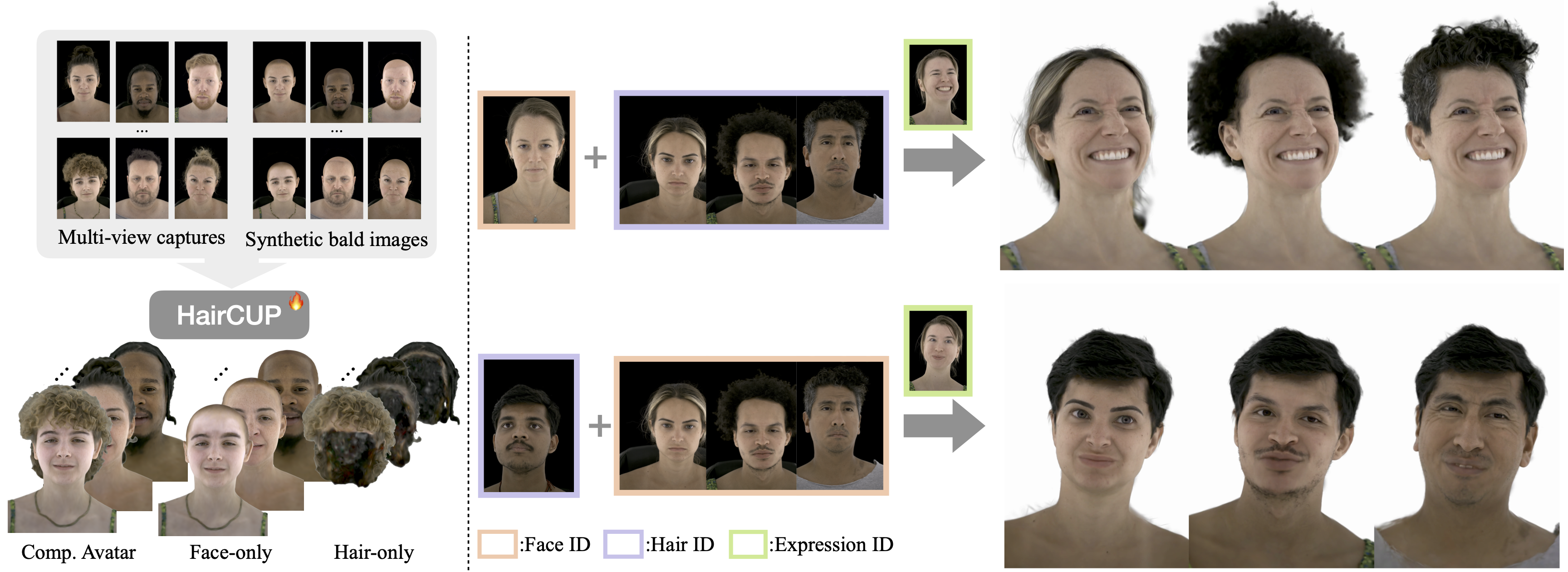}
        \captionof{figure}
        {\textbf{HairCUP}. HairCUP is a compositional universal avatar model that generates relightable Gaussian codec avatars for multiple subjects using a single model. It separately models hair and face, enabling seamless hairstyle transfer between avatars without requiring additional scale optimization or suffering from unnatural hair boundary artifacts.}
        \label{fig:teaser}
    \end{minipage}
\end{strip}

\begin{abstract}
We present a universal prior model for 3D head avatars with explicit hair compositionality.
Existing approaches to build generalizable priors for 3D head avatars often adopt a holistic modeling approach, treating the face and hair as an inseparable entity. This overlooks the inherent compositionality of the human head, making it difficult for the model to naturally disentangle face and hair representations, especially when the dataset is limited. Furthermore, such holistic models struggle to support applications like 3D face and hairstyle swapping in a flexible and controllable manner.
To address these challenges, we introduce a prior model that explicitly accounts for the compositionality of face and hair, learning their latent spaces separately. A key enabler of this approach is our synthetic hairless data creation pipeline, which removes hair from studio-captured datasets using estimated hairless geometry and texture derived from a diffusion prior. By leveraging a paired dataset of hair and hairless captures, we train disentangled prior models for face and hair, incorporating compositionality as an inductive bias to facilitate effective separation.
Our model's inherent compositionality enables seamless transfer of face and hair components between avatars while preserving identity. Additionally, we demonstrate that our model can be fine-tuned in a few-shot manner using monocular captures to create high-fidelity, hair-compositional 3D head avatars for unseen subjects. These capabilities highlight the practical applicability of our approach in real-world scenarios, paving the way for flexible and expressive 3D avatar generation.

\end{abstract}

\section{Introduction}
\label{sec:intro}

Hairstyles are a key aspect of personal identity, reflecting individual style and character.  
Today, people can adopt any hairstyle regardless of ethnicity or natural hair type, making hair one of the most easily transformable physical traits.  
Recent advances in image generation and editing enable people to explore different hairstyles in photos, driving interest in realistic and controllable hair synthesis.  

However, for 3D avatars, achieving comparable hairstyle transfer or editing remains challenging.
Recent developments in 3D avatar technology have brought substantial improvements in visual quality, allowing for disentanglement and independent control over features like expression, view direction, and even lighting~\cite{lombardi2021mvp,zhuang2022mofanerf,cao2022authentic,qian2024gaussianavatars,saito2024relightable,guo2025vid2avatar}. Some studies have introduced generalizable prior models trained on capture data from multiple subjects~\cite{buhler2023preface,yang2024vrmm,li2024uravatar,zheng2024headgap,yu2024one2avatar} or synthetic data~\cite{zielonka2025synshot,saunders2024gasp}, building a latent space of plausible 3D head avatars. These models can then be fine-tuned to create personalized 3D avatars of new subjects, providing significant flexibility.
Despite these advancements, many approaches still rely on holistic modeling that overlooks the inherent compositionality of the head and hair. This limitation makes tasks like hairstyle transfer particularly nontrivial.
While some studies propose methods to reconstruct compositional 3D avatars from video~\cite{Feng2022scarf,Feng2023DELTA} or multi-view capture~\cite{wang2025mega} using hybrid representation of parametric face meshes~\cite{li2017flame,pavlakos2019smplx} and neural volumetric field~\cite{mildenhall2020nerf} or 3D Gaussians~\cite{kerbl20233dgs} for hair, these approaches rely on specialized rendering schemes and require per-subject optimization, lacking a universal prior for scalable hairstyle transfer.

One of the major challenges in learning a disentangled latent space of face and hair from visual data lies in the availability of paired datasets that capture individuals both with and without hair in multi-view setups. Without such data, it becomes difficult to separate hair from facial attributes, as the model lacks a reference for how each person appears without hair. These paired examples provide essential supervision for disentangling face and hair features.
However, collecting them is inherently difficult. Participants are often unwilling to undergo hairless captures, which may require shaving their heads or wearing full hair-concealing caps~\cite{yang2020facescape}. As a result, it remains challenging to obtain diverse, high-quality data for disentanglement.

To overcome these limitations, 
we propose a method to build a 3D hair-compositional prior model for 3D Gaussian head avatars, leveraging studio-captured multi-view data.
To address the absence of paired captures with hair and without hair, we generate synthetic hairless data. To get multiview-consistent hairless data, we register a bald mesh for each subject and get hairless texture using a diffusion prior. During training, we mask out the hair region from images and render the bald mesh to synthesize hairless images. We then extend the holistic prior model~\cite{li2024uravatar} to learn a compositional 3D prior that disentangles face and hair. Leveraging the synthetic paired dataset, our model learns disentangled latent spaces of face and hair in a data-efficient manner.
This universal model enables face and hair transfer across training identities by conditioning each component on separate identities. Furthermore, it can be fine-tuned on unseen captures to generate personalized 3D avatars with independent control over face and hair components.
In summary, our contributions are:
\begin{itemize}
    \item A 3D-consistent synthetic hairless data generation pipeline for creating paired datasets, enabling effective face-hair disentanglement.
    \item A compositional universal model that independently represents 3D hairless heads and hair, allowing explicit control over facial appearance and hairstyles.
    \item Using our universal avatar model as a prior, we adapt to novel identities through fine-tuning, enabling hairstyle transfer and dehairing on unseen subjects.
\end{itemize}

\section{Related Work}
\label{sec:related}

\subsection{3D Animatable Head Avatar}
\label{sec:related_avatar}

Recent advancements in 3D animatable head avatars have evolved from mesh-based parametric models~\cite{blanz19993dmm,li2017flame,cao2013facewarehouse,tran2018nonlinear} to neural rendering techniques~\cite{thies2019deferred}, which enhance realism by combining facial meshes with neural textures~\cite{khakhulin2022rome,ma2021pica,grassal2022nha,cao2022authentic}.  
Neural implicit representations~\cite{sitzmann2019deepvoxels,mildenhall2020nerf,mescheder2019occupancy} further revolutionized volumetric avatar synthesis, enabling high-quality novel view synthesis and expression-driven animations~\cite{zhuang2022mofanerf,zheng2022imavatar,gafni2021nerface,gao2022nerfblenshape,bai2023monoavatar,zielonka2023insta}. To address the high computational costs of volumetric rendering, point-based methods~\cite{zheng2023pointavatar} and UV-anchored local fields~\cite{lombardi2021mvp,yang2023travatar,bai2024monoavatar++} provide efficient alternatives for real-time applications.  
Recently, 3D Gaussian representations~\cite{kerbl20233dgs} have emerged as a powerful alternative, demonstrating high-fidelity yet memory-efficient avatars~\cite{saito2024relightable,qian2024gaussianavatars,xu2023gaussianheadavatar,giebenhain2024npga,xiang2024flashavatar}.  
Additionally, generalizable prior models trained on multi-subject capture~\cite{buhler2023preface,yang2024vrmm,li2024uravatar,zheng2024headgap,yu2024one2avatar,liu2025lucas,guo2025vid2avatar} or synthetic data~\cite{zielonka2025synshot,saunders2024gasp} enable novel identity reconstruction without requiring extensive per-subject optimization. However, most approaches treat the face and hair as a unified entity. Our method builds a compositional prior for 3D Gaussian avatars, disentangling face and hair to facilitate tasks like hairstyle transfer.

\subsection{Compositional Human Modeling}
\label{sec:related_comp}
Compositional modeling decomposes human representations into distinct elements such as face, hair, and clothing, allowing independent control and flexible editing~\cite{natsume2018rsgan,shi2022semanticstylegan,kim2024gala,cha2024pegasus,cha2024perse,xiang2021modeling,li2023megane}. This approach enables applications like hairstyle and garment transfer~\cite{kim2023ncho,Feng2023DELTA,Feng2022scarf,kim2024gala,lin2024layga,wang2025mega}.  
Recent studies extend compositionality to 3D avatars. PEGASUS and PERSE~\cite{cha2024pegasus,cha2024perse} introduce disentangled facial attributes for personalized avatar generation, while DELTA~\cite{Feng2023DELTA} and TECA~\cite{zhang2024teca} combine the SMPL-X~\cite{pavlakos2019smplx} for body and face with volumetric rendering~\cite{mildenhall2020nerf,muller2022instantngp} for clothing and hair. However, mesh-based methods inherently limit facial resolution, and hair dynamics rely on linear blend skinning (LBS) deformations. 
MeGA~\cite{wang2025mega} proposes a hybrid design combining a face mesh with neural textures and 3D Gaussians for hair. While this design improves rendering fidelity, it introduces additional training complexity due to a visibility test between hair Gaussians and the face mesh.
Our model directly builds upon fully Gaussian-based avatars~\cite{qian2024gaussianavatars,xu2023gaussianheadavatar,xiang2024flashavatar,saito2024relightable,li2024uravatar}, learning disentangled latent spaces for facial expressions and hair motion, which enables high-fidelity compositional avatars with enhanced realism.

\subsection{3D Hair Modeling}
\label{sec:related_hair}
Hair is challenging to model due to its fine-scale geometry and complex photometric behavior. Since parametric head models (\eg FLAME~\cite{li2017flame}) omit hair, most approaches treat it as a separate component, represented by explicit strands~\cite{paris2004capture,nam2019strand,rosu2022neuralstrands,sklyarova2023neuralhaircut}, neural volumetric fields~\cite{saito20183d,wang2023neuwigs,Feng2023DELTA,zhang2024teca}, or 3D Gaussian primitives~\cite{zakharov2024gshaircut,luo2024gaussianhair,wang2025mega}. 
Strand-based methods remain prevalent due to their compatibility with standard rendering and simulation pipelines. These methods typically estimate 3D strands by leveraging 2D hair orientation cues from multi-view images~\cite{paris2004capture,nam2019strand,rosu2022neuralstrands,sklyarova2023neuralhaircut}. NeuralHaircut~\cite{sklyarova2023neuralhaircut} improves strand reconstruction using a diffusion-based hairstyle prior~\cite{ho2020ddpm}, while HAAR~\cite{sklyarova2024haar} and DiffLocks~\cite{rosu2025difflocks} employ generative diffusion models to synthesize 3D strands from text or a single image.
GaussianHair~\cite{luo2024gaussianhair} proposes aligning 3D Gaussians~\cite{kerbl20233dgs} along reconstructed strands for view-consistent, relightable rendering. 
Our method also adopts a Gaussian-based representation, but focuses on building a universal prior by anchoring 3D Gaussians on the UV map of a shared-topology facial mesh, enabling consistent hair parameterization across subjects.

\begin{figure*}[t]
\centering
\includegraphics[width=0.90\linewidth, trim={0cm 0cm 1cm 1cm}]{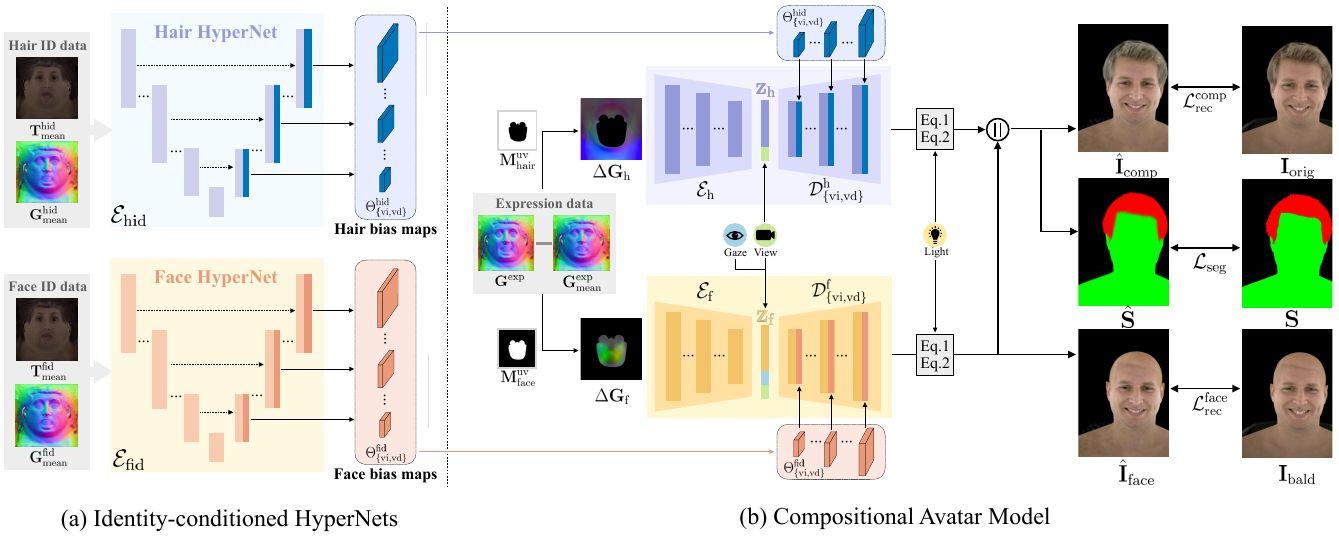}
\captionof{figure}{\textbf{Model overview.} HairCUP comprises ID-conditioned face/hair hypernetworks and a compositional avatar model.  
(a) The hypernetworks take UV-unwrapped mean albedo and geometry maps as input, generating multi-scale bias maps as ID conditions, added to each layer of the face/hair Gaussian decoders. 
(b) The compositional avatar model consists of a hair motion and face expression encoder $\mathcal{E}_{\{\mathrm{f,h}\}}$ and Gaussian decoders $\mathcal{D}^{\{\mathrm{f,h}\}}_{\{\mathrm{vi,vd}\}}$. The encoders produce hair motion code $\vec{z}_{\mathrm{h}}$ and expression code $\vec{z}_{\mathrm{f}}$ from the delta geometry map $\Delta\mat{G}^{\mathrm{exp}}$, which the decoders use to generate Gaussians.  
During training, face/hair ID and expression data come from the same subject, forming a 3D avatar supervised by multi-view capture data ($\mathcal{L}_{\mathrm{rec}}^{\mathrm{comp}}$). One-hot labels assigned to face/hair Gaussians are rendered into segmentation maps and supervised by face/hair masks for disentanglement ($\mathcal{L}_{\mathrm{seg}}$). Additionally, face Gaussians are separately rendered and supervised with synthetic bald data ($\mathcal{L}_{\mathrm{rec}}^{\mathrm{face}}$).  
At test time, face/hair ID, and expression data can be mixed across subjects.}
\label{fig:pipeline}
\end{figure*}

\section{Method}
\label{sec:method}
\subsection{Preliminaries: Religtable 3D Gaussians}
\label{subsection:preliminaries}
Our approach builds on URAvatar~\cite{li2024uravatar}, a universal 3D avatar model that extends person-specific relightable 3D Gaussian avatars~\cite{saito2024relightable} to multiple subjects. We first introduce its core component: relightable 3D Gaussians. More details of URAvatar are provided in the Supp. Mat.

Given multi-view studio-captured images with time-multiplexed lighting and facial tracked meshes, \citet{saito2024relightable} build a relightable 3D head avatar model. While following the geometric parameterization of 3D Gaussians originally proposed by \citet{kerbl20233dgs}, they advance the appearance model based on learnable radiance transfer for relightability.
The outgoing radiance of each Gaussian $\vec{c}_{k}$ can be computed by summation of diffuse color $\vec{c}_{k}^{\mathrm{diffuse}}$ and specular reflection $\vec{c}_{k}^{\mathrm{specular}}$, 
where each term can be efficiently evaluated. The diffuse color is computed as:
\begin{equation}
\label{eq:diffuse}
\vec{c}_{k}^{\mathrm{diffuse}}
=
\vec{\rho}_{k} \odot
\sum_{i=1}^{(n+1)^2}\mat{L}_{i} \odot
\vec{d}_{k}^{i},
\end{equation}
where $\vec{\rho}_{k}$ is a static albedo, $\mat{L}_{i}$ and $\vec{d}_{k}^{i}$ are the $i$-th element of $n$-th order spherical harmonics(SH) of the incident lights and learnable transfer function, respectively. 
The specular term is formulated based on \citet{wang2009all}:
\begin{align}
\label{eq:specular}
\vec{c}_{k}^{\mathrm{specular}}(\vec{\omega}_{k}^{o}) &=
v_{k}(\omega_{k}^{o})
\int_{\mathbb{S}^2}
\mat{L}(\vec{\omega})
G_{s}(\vec{\omega};\vec{q}_{k}, \sigma_{k}) \, d\vec{\omega}, \\
\label{eq:q_vector}
\vec{q}_{k} &= 2(\vec{\omega}_{k}^{o} \cdot \vec{n}_{k})\vec{n}_{k} - 
\vec{\omega}_{k}^{o},
\end{align}
where $\vec{\omega}_{k}^{o}$ is a viewing direction at the Gaussian center, $\vec{v}_{k}(\vec{\omega}_{k}^{o})$ is view-dependent visibility, $G_{s}(\vec{\omega};\vec{q}_{k}, \sigma_{k})$ is a spherical Gaussian~\cite{wang2009all} with the lobe axis $\vec{q}_{k}$ and roughness $\sigma_{k}$, and $n_{k}$ denotes a view-dependent surface normal.

\subsection{Compositional Universal Avatar Model}
\label{subsection:composition}
Our hair-compositional 3D head avatar model produces separate sets of relightable 3D Gaussians for face and hair from input expression code and hair motion code from encoders.
To build a compositional universal prior model for 3D head avatars, our model consists of a bald head geometry decoder, three sets of identity-conditioned hypernetworks~\cite{cao2022authentic,li2024uravatar} and 3D Gaussian decoders for hair, face, and eyes respectively. \Cref{fig:pipeline} shows an overview of our hair-compositional universal model for 3D Gaussian avatars.
\paragraph{Expression/hair motion encoder.}
Our model employs separate encoders for facial expression and hair motion, learning their respective latent spaces. Following URAvatar~\cite{li2024uravatar}, the encoders take the delta of geometry maps, UV-mapped geometry of the tracked meshes, computed as $\Delta \mat{G}^{\mathrm{exp}}=\mat{G}^{\mathrm{exp}}-\mat{G}_{\mathrm{mean}}^{\mathrm{exp}}$, where $\mat{G}^{\mathrm{exp}}$ is the current frame’s geometry map and $\mat{G}_{\mathrm{mean}}^{\mathrm{exp}}$ is the mean over all frames. To decouple expression from hair motion, we use predefined face and hair region masks, $\mat{M}_\mathrm{f}$ and $\mat{M}_\mathrm{h}=1-\mat{M}_\mathrm{f}$ in UV space. The expression encoder $\mathcal{E}_\mathrm{f}$ and hair motion encoder $\mathcal{E}_\mathrm{h}$ process masked delta geometry and output the mean and covariance of their respective latent codes:
\begin{equation}
    \label{eq:encoder}
    \vec{\mu}_{\{ \mathrm{f,h} \}}, \vec{\sigma}_{\{ \mathrm{f,h} \}} =
    \mathcal{E}_{\{ \mathrm{f,h} \}}(
        \Delta \mat{G}_{\{ \mathrm{f,h} \}}; \Theta_{\{ \mathrm{f,h} \}}
    ),
\end{equation}
where $\Delta \mat{G}_{\{\mathrm{f,g}\}} =\mat{M}_\mathrm{\{\mathrm{f,g}\}}^{\mathrm{uv}} \odot \Delta \mat{G}^{\mathrm{exp}}$. Then, an expression code $\vec{z}_\mathrm{f}$ and hair motion code $\vec{z}_\mathrm{h}$ are sampled based on reparameterization trick~\cite{kingma2014vae}:
\begin{equation}
    \label{eq:sample_latent}
    \vec{z}_\mathrm{\left\{f, h\right\}}=
    \vec{\mu}_\mathrm{\left\{f, h\right\}} +
    \vec{\sigma}_\mathrm{\left\{f, h\right\}} \cdot
    \mathcal{N}(\vec{0}, \mat{I}),
\end{equation}
where $\mathcal{N}(\vec{0}, \mat{I})$ is a standard normal distribution. The expression code and hair motion code are fed into the face Gaussian decoder and hair Gaussian decoder to produce 3D Gaussians based on the current state of the face and hair.
\paragraph{Identity-conditioned hypernetworks.}
Building on the identity-conditioned hypernetwork~\cite{cao2022authentic, li2024uravatar}, which encodes identity information to guide avatar appearance, our model introduces separate hypernetworks $\mathcal{E}_{\{\mathrm{fid, hid}\}}$ for face and hair, with corresponding network parameters $\Phi_{\{\mathrm{fid, hid}\}}$:
\begin{gather}
\label{eq:hyper_face}
\Theta^{\mathrm{fid}}_{\mathrm{vi}},
\Theta^{\mathrm{fid}}_{\mathrm{vd}},
\{o_{k}^\mathrm{f}, \vec{\rho}_{k}^\mathrm{f}\}^{N_\mathrm{f}}_{k=1} =
\mathcal{E}_{\mathrm{fid}}(\mat{T}_{\mathrm{mean}}^{\mathrm{fid}}, \mat{G}_{\mathrm{mean}}^{\mathrm{fid}};
\Phi_{\mathrm{fid}}), \\
\label{eq:hyper_hair}
\hspace{-3.9mm}
\Theta^{\mathrm{hid}}_{\mathrm{vi}},
\Theta^{\mathrm{hid}}_{\mathrm{vd}},
\{o_{k}^\mathrm{h}, \vec{\rho}_{k}^\mathrm{h}\}^{N_\mathrm{h}}_{k=1} =
\mathcal{E}_{\mathrm{hid}}(\mat{T}_{\mathrm{mean}}^{\mathrm{hid}}, \mat{G}_{\mathrm{mean}}^{\mathrm{hid}};
\Phi_{\mathrm{hid}}),
\end{gather}
where $\mat{T}_\mathrm{mean}^{\{\mathrm{fid,hid}\}}, \mat{G}_\mathrm{mean}^{\{\mathrm{fid,hid}\}}$ are the mean albedo texture map and mean geometry map for face and hair, 
$\Theta^{\mathrm{\{fid, hid\}}}_{\mathrm{vi}},
\Theta^{\mathrm{\{fid, hid\}}}_{\mathrm{vd}}$ are the bias maps injected to face and hair Gaussian decoders, $N_{\{\mathrm{f,h}\}}$ are the number of face and hair Gaussians, and $o_{k}^\mathrm{\{f,h\}}, \vec{\rho}_{k}^\mathrm{\{f,h\}}$ are the expression-agnostic opacity and albedo color of face and hair Gaussians following URAvatar~\cite{li2024uravatar}.
These bias maps work as an identity conditioning mechanism for our universal model to serve as a person-specific avatar model based on the input mean albedo texture map $\mat{T}_\mathrm{mean}^{\{\mathrm{fid,hid}\}}$ and geometry map $\mat{G}_\mathrm{mean}^{\{\mathrm{fid,hid}\}}$. 
We use the same eye hypernetwork from URAvatar~\cite{li2024uravatar} to produce bias maps for the eye decoders.
\paragraph{Bald geometry prediction.}
Unlike \citet{saito2024relightable} and URAvatar~\cite{li2024uravatar}, which have a per-frame geometry decoder, we have a separate encoder $\mathcal{E}_\mathrm{b}$ and decoder $\mathcal{D}_\mathrm{b}$ which predicts vertex positions of the static bald tracked mesh from the mean geometry map $\mat{G}_{\mathrm{mean}}^{\mathrm{fid}}$ as follows:
\begin{align}
\label{eq:enc_bald}
    \vec{z}_{\mathrm{b}} &= 
    \mathcal{E}_\mathrm{b}(\mat{G}_{\mathrm{mean}}^{\mathrm{fid}};\Theta_\mathrm{b}), \\
\label{eq:dec_bald}
    \hat{\mat{V}}_{\mathrm{b}}&=
    \mathcal{D}_\mathrm{b}(\vec{z}_\mathrm{b};\Phi_\mathrm{b}),
\end{align}
where $\vec{z}_\mathrm{b}$ is the bald geometry latent code and $\hat{\mat{V}}_{\mathrm{b}}$ is the predicted vertex positions based on the shared topology of tracked meshes. The unwrapped bald geometry map $\hat{\mat{G}}_{\mathrm{b}}$ and mean geometry map $\mat{G}_{\mathrm{mean}}^{\mathrm{fid}}$ are combined to preserve the face region using the UV face region mask $\mat{M}_{\mathrm{face}}^{\mathrm{uv}}$ as $\mat{G}_{\mathrm{b}}=\mat{M}_{\mathrm{face}}^{\mathrm{uv}} \odot \mat{G}_{\mathrm{mean}}^{\mathrm{fid}} + (1-\mat{M}_{\mathrm{face}}^{\mathrm{uv}}) \odot \hat{\mat{G}}_{\mathrm{b}}$. The combined bald geometry map $\mat{G}_{\mathrm{b}}$ anchors 3D Gaussian primitives, defining Gaussian translations relative to the mesh. Since hair originates from the scalp, modeling hair relative to the bald head geometry enables natural hairstyle transfer across avatars with varying head shapes and sizes.
\paragraph{Face/hair Gaussian decoders.}
To generate 3D Gaussians for face and hair in a disentangled manner, we use separate decoders for each component, where the Gaussians are parameterized on the UV map of the tracked mesh. Specifically, we employ view-independent decoders $\mathcal{D}_{\mathrm{vi}}^{\{\mathrm{f,h}}\}$ for geometric attributes and view-dependent decoders $\mathcal{D}_{\mathrm{vd}}^{\{\mathrm{f,h}\}}$ for appearance attributes, following URAvatar~\cite{li2024uravatar}.
The decoders $\mathcal{D}_{\mathrm{vi}}^{\{\mathrm{f,h}\}}$ predict:
\begin{equation}
\hspace{-3mm}
    \label{eq:dec_face_gs_vi}
    \{
    \delta \vec{t}_{k}^{\mathrm{f}}, 
    \vec{q}_{k}^{\mathrm{f}}, \vec{s}_{k}^{\mathrm{f}}, 
    \vec{d}_{k}^{\mathrm{f}},
    \sigma_{k}^{\mathrm{f}}
    \}_{k=1}^{N_{\mathrm{f}}}
    =
    \mathcal{D}_{\mathrm{vi}}^{\mathrm{f}}(
    \vec{z}_{\mathrm{f}}, 
    \vec{e}_{\{\mathrm{l, r}\}};
    \Theta_{\mathrm{vi}}^{\mathrm{fid}}, \Phi_{\mathrm{vi}}^{\mathrm{f}}),
\end{equation}
\begin{equation}
    \label{eq:dec_hair_gs_vi}
    \{
    \delta \vec{t}_{k}^{\mathrm{h}}, 
    \vec{q}_{k}^{\mathrm{h}}, \vec{s}_{k}^{\mathrm{h}}, 
    \vec{d}_{k}^{\mathrm{h}},
    \sigma_{k}^{\mathrm{h}}
    \}_{k=1}^{N_{\mathrm{h}}}
    =
    \mathcal{D}_{\mathrm{vi}}^{\mathrm{h}}(
    \vec{z}_{\mathrm{h}} ;
    \Theta_{\mathrm{vi}}^{\mathrm{hid}}, \Phi_{\mathrm{vi}}^{\mathrm{h}}),
\end{equation}
where $\delta \vec{t}_{k}$ is the position offset relative to the bald mesh, and $\vec{q}_{k}, \vec{s}_{k}, \vec{d}_{k}, \sigma_{k}$ denote the orientation, scale, SH coefficients for color and monochrome components, and roughness parameter~\cite{saito2024relightable}.
The decoders $\mathcal{D}_{\mathrm{vd}}^{\{\mathrm{f,h}\}}$ predict:
\begin{equation}
    \label{eq:dec_face_gs_vd}
    \{
    \delta \vec{n}_{k}^{\mathrm{f}}, 
    v_{k}^{\mathrm{f}}
    \}_{k=1}^{N_{\mathrm{f}}}
    =
    \mathcal{D}_{\mathrm{vd}}^{\mathrm{f}}(
    \vec{z}_{\mathrm{f}}, 
    \vec{e}_{\{\mathrm{l, r}\}},
    \vec{\omega}_{o};
    \Theta_{\mathrm{vd}}^{\mathrm{fid}}, \Phi_{\mathrm{vd}}^{\mathrm{f}}),
\end{equation}
\begin{equation}
    \label{eq:dec_hair_gs_vd}
    \{
    \delta \vec{n}_{k}^{\mathrm{h}}, 
    v_{k}^{\mathrm{h}}
    \}_{k=1}^{N_{\mathrm{h}}}
    =
    \mathcal{D}_{\mathrm{vd}}^{\mathrm{h}}(
    \vec{z}_{\mathrm{h}},
    \vec{\omega}_{o};
    \Theta_{\mathrm{vd}}^{\mathrm{hid}}, \Phi_{\mathrm{vd}}^{\mathrm{h}}),
\end{equation}
where $\delta \vec{n}_{k}$ is the view-dependent normal offset, and $v_{k}$ represents the learned visibility term. Unlike the face decoder, the hair decoder does not take eye gaze direction $\vec{e}_{\{\mathrm{l, r}\}}$ as input, as it does not influence the hair state.
The predicted delta translations $\delta \vec{t}_{k}^{\{\mathrm{f,h}\}}$ are then added to the corresponding texel $\vec{t}_{k}^{\mathrm{b}}$ of the bald geometry map $\mat{G}_{\mathrm{b}}$:
\begin{equation}
\label{eq:add_trans}
    \vec{t}_{k}^{\{\mathrm{f,h}\}}
    =
    \vec{t}_{k}^{\mathrm{b}}
    + \delta \vec{t}_{k}^{\{\mathrm{f,h}\}}.
\end{equation}
The face Gaussian’s delta normal $\delta \vec{n}_{k}^{\mathrm{f}}$ is added to the mesh normal $\vec{n}_{k}^{\mathrm{b}}$ derived from the bald geometry map $\mat{G}_{\mathrm{b}}$, while the hair Gaussian’s delta normal $\delta \vec{n}_{k}^{\mathrm{h}}$ is added to $\vec{n}^{\mathrm{hid}}_{k}$ from the mean geometry map $\mat{G}_{\mathrm{mean}}^{\mathrm{hid}}$, which offers a more consistent geometric reference for hair Gaussians:
\begin{align}
\label{eq:add_face_normal}
    \vec{n}_{k}^{\mathrm{f}}
    &=
    \vec{t}_{k}^{\mathrm{b}}
    + \delta \vec{t}_{k}^{\mathrm{f}} \\
\label{eq:add_hair_normal}
    \vec{n}_{k}^{\mathrm{h}}
    &=
    \vec{n}^{\mathrm{hid}}_{k}
    + \delta \vec{n}_{k}^{\mathrm{h}}.
\end{align}
For eye Gaussian decoders, we follow URAvatar~\cite{li2024uravatar} except the unified specular visibility decoder for simplicity:
\begin{equation}
\hspace{-3mm}
\label{eq:dec_eye_gs_vi}
    \{
    \vec{q}_{k}^{\mathrm{e}}, 
    \vec{s}_{k}^{\mathrm{e}}, 
    o_{k}^{\mathrm{e}}, 
    \vec{d}_{k}^{\mathrm{e}},
    \sigma_{k}^{\mathrm{e}}
    \}_{k=1}^{N_\mathrm{e}}
    =
    \mathcal{D}_{\mathrm{vi}}^{\mathrm{e}}
    (
        \vec{e}_{\{l,r\}};
        \Theta^{\mathrm{e}}_{\mathrm{vi}},
        \Phi^{\mathrm{e}}_{\mathrm{vi}}
    ),
\end{equation}
\begin{equation}
\label{eq:dec_eye_gs_vd}
    \{
        \vec{\rho}_{k}^{\mathrm{e}},
        v_{k}^{\mathrm{e}}
    \}_{k=1}^{N_\mathrm{e}}
    =
    \mathcal{D}_{\mathrm{vd}}^{\mathrm{e}}
    (
        \vec{e}_{\{l,r\}},
        \vec{\omega}_{o};
        \Theta_{\mathrm{vd}}^{\mathrm{e}},
        \Phi_{\mathrm{vd}}^{\mathrm{e}}
    ),
\end{equation}
where $\mathcal{D}_{\mathrm{vi}}^{\mathrm{e}},\mathcal{D}_{\mathrm{vd}}^{\mathrm{e}}$ are view-independent and view-dependent Gaussian decoders with parameters $\Phi_{\mathrm{vi}}^{\mathrm{e}}$ and $\Phi_{\mathrm{vd}}^{\mathrm{e}}$, $\Theta_{\mathrm{vi}}^{\mathrm{e}}$ and $\Theta_{\mathrm{vd}}^{\mathrm{e}}$ are bias maps from the eye hypernetwork $\mathcal{E}_{\mathrm{eye}}$.
Given the incident light $\mat{L}$ and the Gaussian attributes of the hair, face, and eyes, we compute the outgoing radiance of each Gaussian through \cref{eq:diffuse,eq:specular}. Then, these Gaussians can be rendered with a standard Gaussian rasterizer~\cite{kerbl20233dgs}.

\begin{figure}
    \captionsetup[subfigure]{font=footnotesize}
     \centering
     \begin{subfigure}[b]{0.24\columnwidth}
         \centering
         \includegraphics[width=\textwidth]{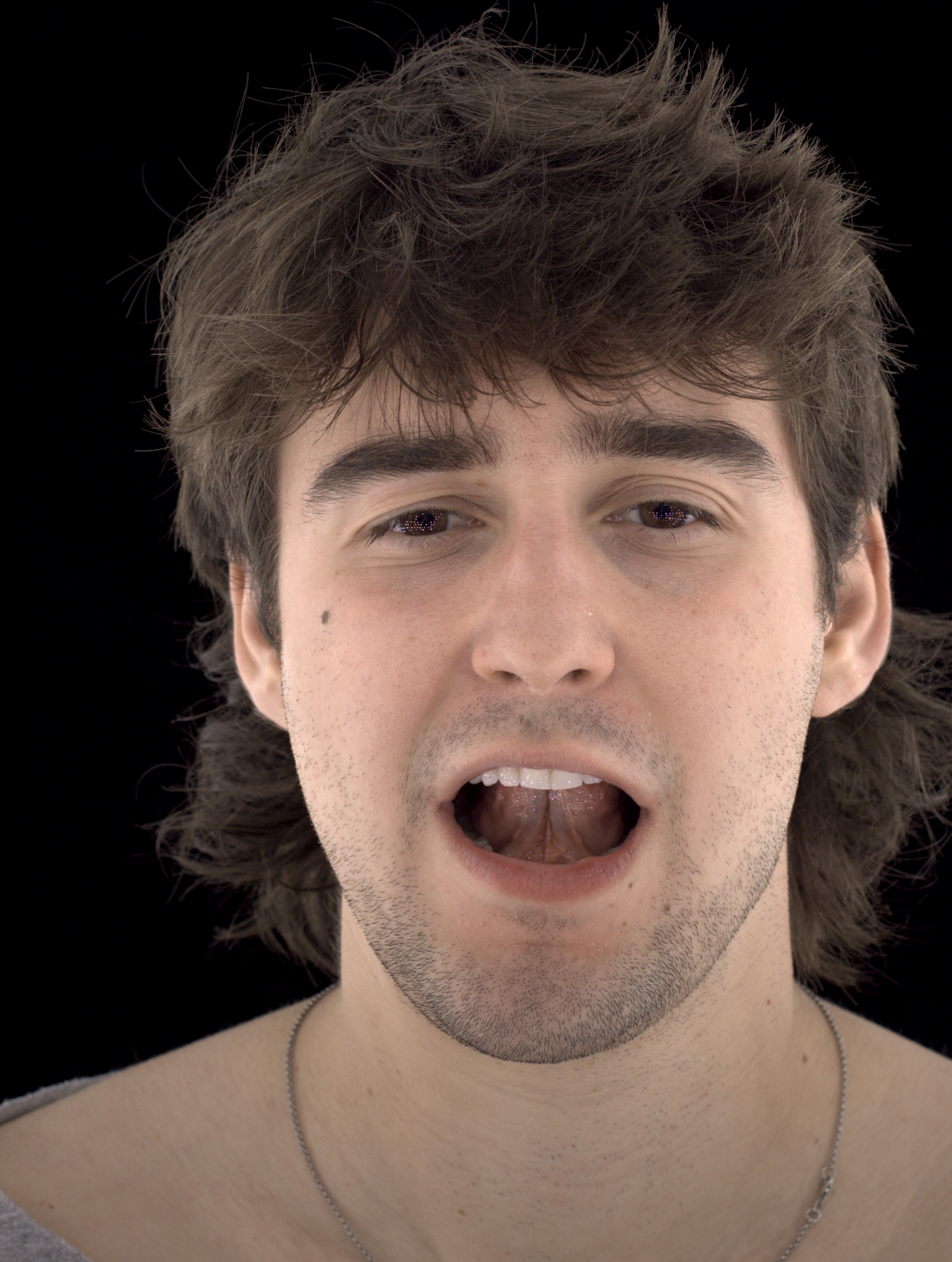}
         \caption{Capture image}
         \label{fig:mugsy_cam_img}
     \end{subfigure}
     \begin{subfigure}[b]{0.24\columnwidth}
         \centering
         \includegraphics[width=\textwidth]{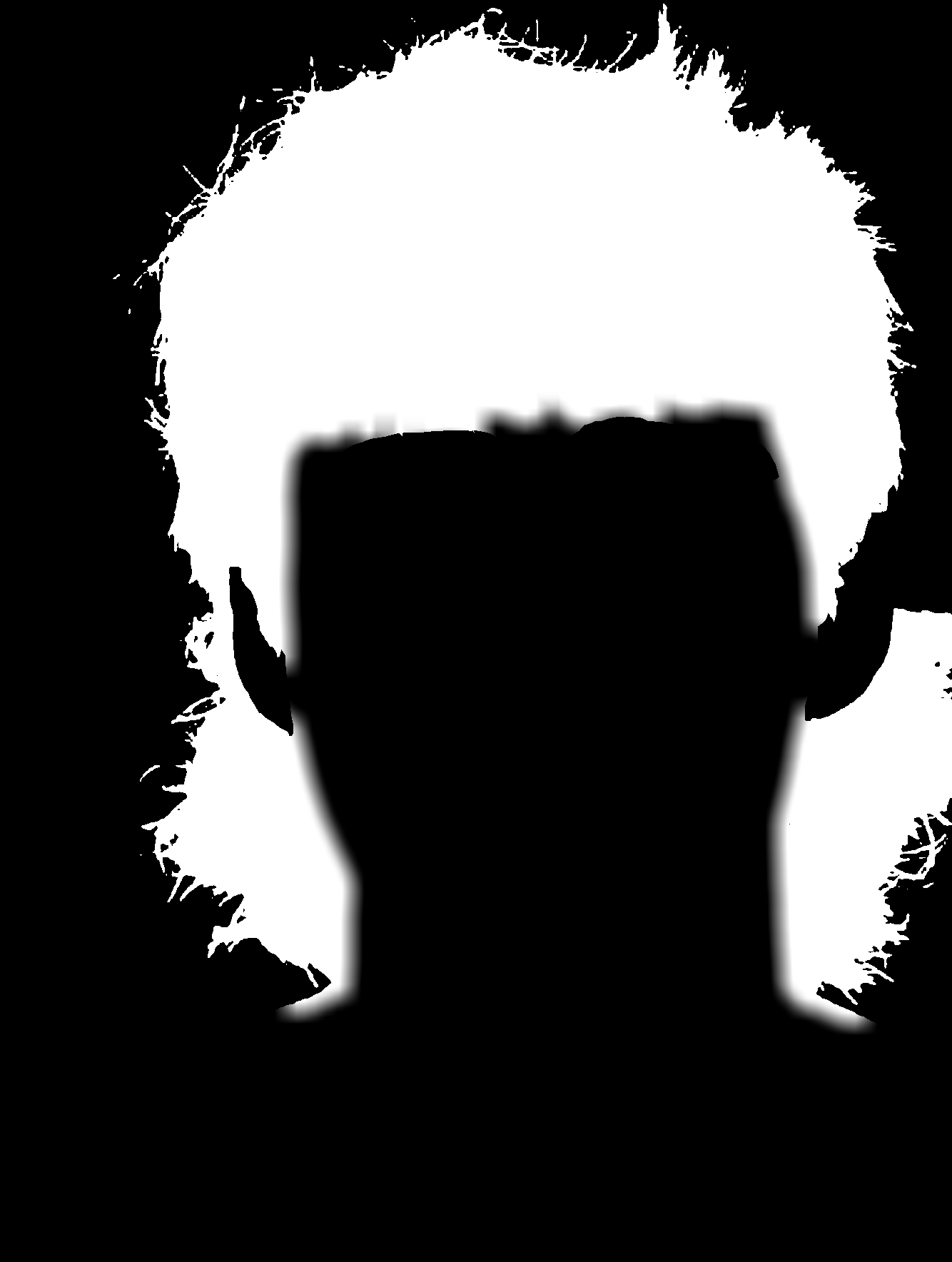}
         \caption{Hair mask}
         \label{fig:hair_mask}
     \end{subfigure}
     \begin{subfigure}[b]{0.24\columnwidth}
         \centering
         \includegraphics[width=\textwidth]{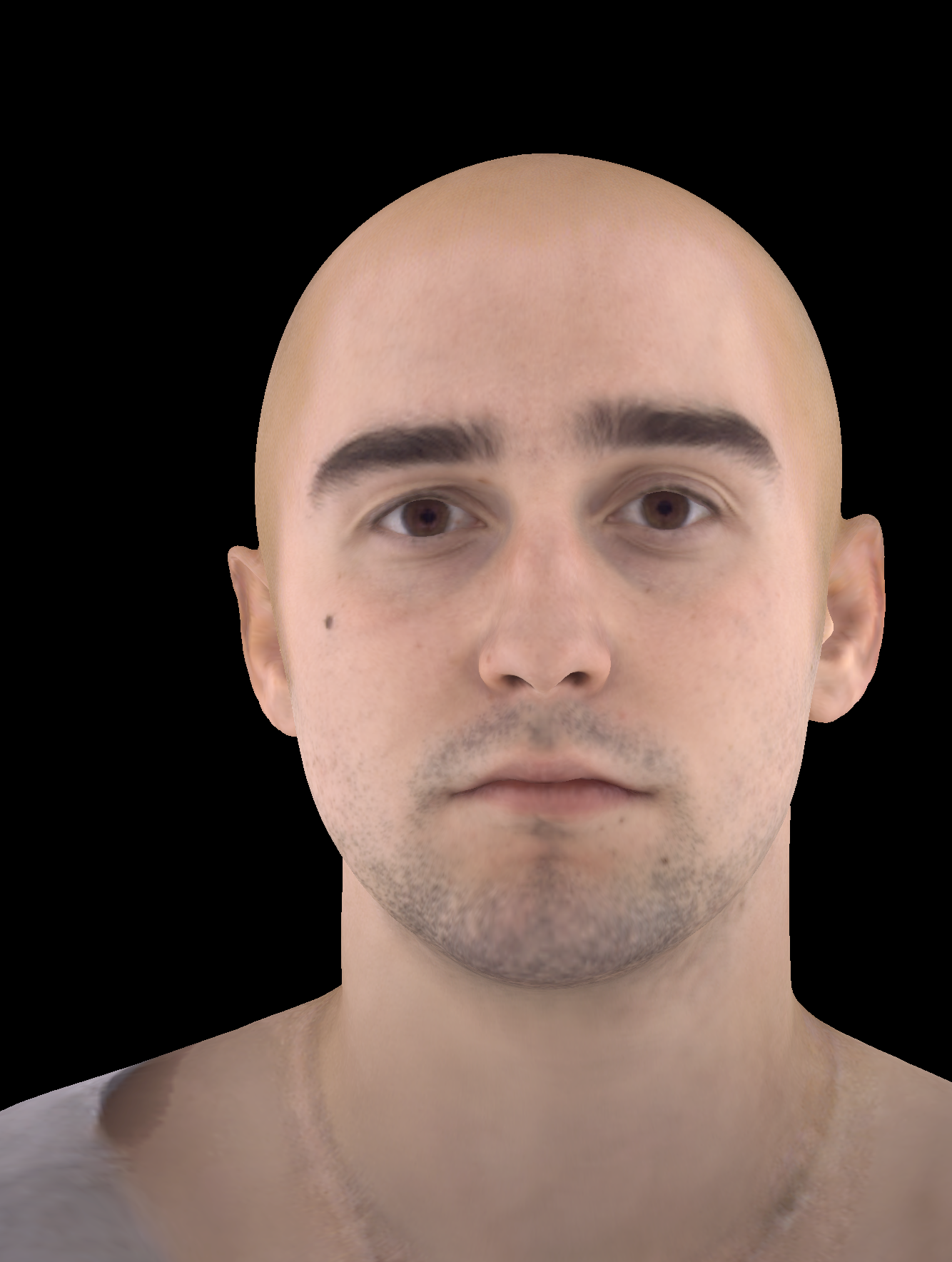}
         \caption{Bald mesh}
         \label{fig:bald_mesh_render}
     \end{subfigure}
     \begin{subfigure}[b]{0.24\columnwidth}
         \centering
         \includegraphics[width=\textwidth]{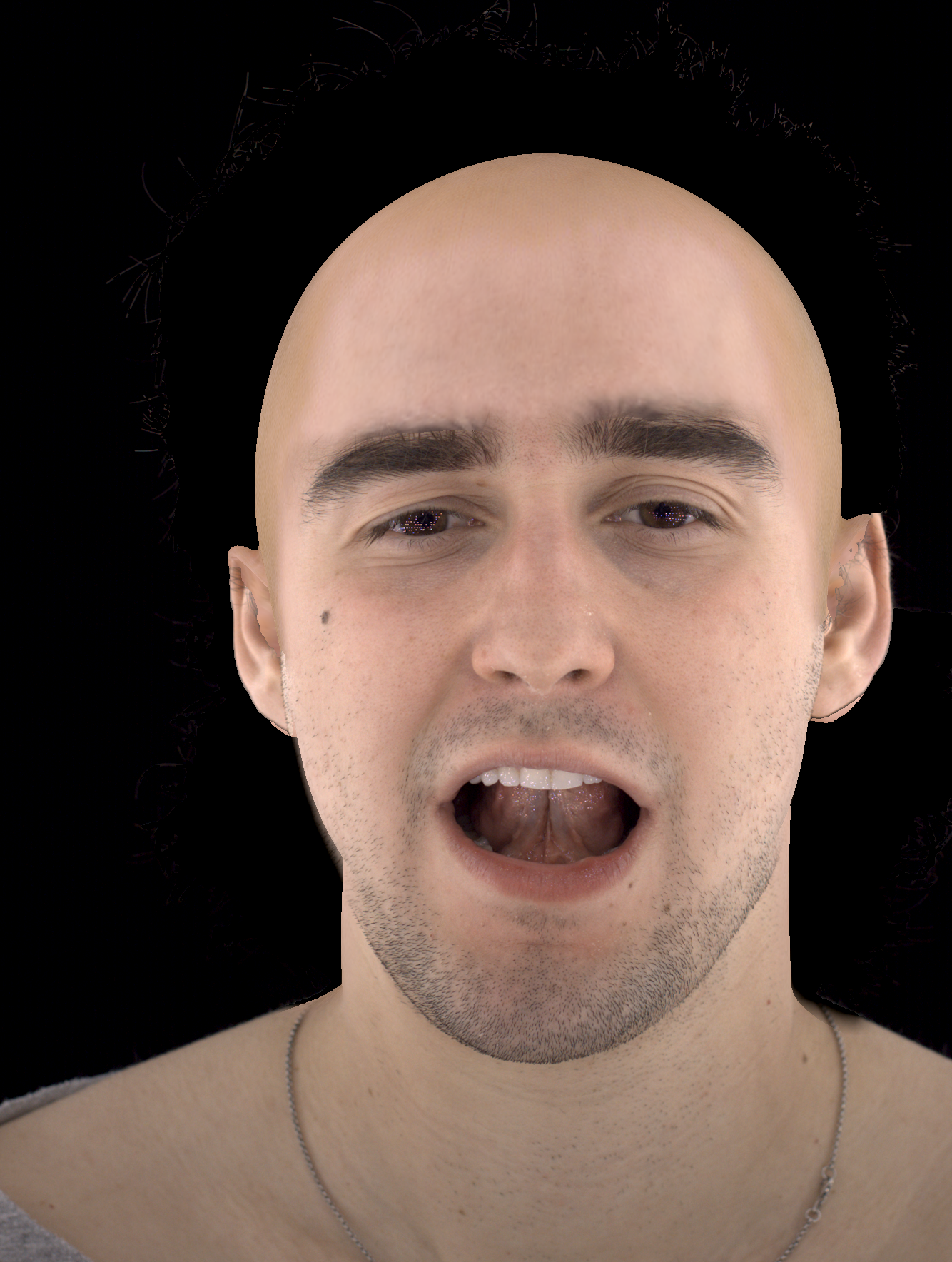}
         \caption{Composition}
         \label{fig:comp_img}
     \end{subfigure}
    \vspace{-3mm}
    \caption{\textbf{Synthetic bald image.} We get synthetic bald images by compositing (a) the original capture with (c) the rendered bald mesh. Hair masks (b) remove the hair region, allowing the bald mesh to be rendered into (d) the final composited image.} 
\label{fig:pseudo_bald_img}
\end{figure}

\subsection{Synthetic Bald Image Generation}
\label{subsection:baldify}
Unlike previous work that performs per-image dehairing~\cite{wu2022hairmapper,cha2024pegasus,zhang2024stablehair} or estimates the scalp region using an average skin color~\cite{Feng2023DELTA}, we process a face tracking mesh from a neutral expression frame to obtain a bald counterpart with the same topology. We then optimize a UV texture map of the bald mesh using a color reconstruction loss for the visible face region and a Score Distillation Sampling (SDS) loss~\cite{poole2023dreamfusion} to complete the occluded scalp region with diffusion priors~\cite{lee2024gtu,weber2024nerfiller,wu2024reconfusion,kim2024gala}.
To generate paired haired and hairless images across different expressions, we remove hair regions from haired images using segmentation masks and render the bald mesh onto the masked regions, as shown in \cref{fig:pseudo_bald_img}.
\paragraph{Texture map parameterization.}
We model the UV texture of the bald mesh with a coordinate-based MLP~\cite{sitzmann2019siren} that takes UV coordinates and view directions as inputs to capture view-dependent texture variations. The view-dependent texture map $\mat{T}_\mathrm{bald}\in \mathbb{R}^{H \times W \times 3}$ is computed as:
\begin{equation}
\label{eq:tex_mlp}
    \mat{T}_\mathrm{bald}(i,j)=f_{\theta}((\gamma(u, v),\gamma_\mathrm{SH}(\vec{d}))),
\end{equation}
where $(i, j) \in [0, H-1] \times [0, W-1]$ are texture map pixel coordinates, $u, v \in [0,1]$ are normalized UV coordinates, $\vec{d} \in \mathbb{R}^{3}$ is the view direction, and $\gamma(\cdot), \gamma_\mathrm{SH}(\cdot)$ denote positional and spherical harmonics encodings.
\paragraph{Bald texture optimization.}
We optimize the texture MLP using:
(1) Reconstruction loss from multi-view images of the bald mesh's registered frame and
(2) SDS loss~\cite{poole2023dreamfusion} to hallucinate the occluded scalp region with diffusion prior:
\begin{equation}
\label{eq:bald_optim}
    \mathcal{L}_\mathrm{bald}=
    \lambda_\mathrm{bald}^\mathrm{rec} 
    \mathcal{L}_\mathrm{bald}^\mathrm{rec} + 
    \lambda_\mathrm{bald}^\mathrm{sds} \mathcal{L}_\mathrm{bald}^\mathrm{sds},
\end{equation}
where $\mathcal{L}_{\mathrm{bald}}^\mathrm{rec}$ is L1 loss over the face region $\hat{\mat{M}}_{\mathrm{face}}$, and $\mathcal{L}_\mathrm{bald}^\mathrm{sds}$ is SDS loss. The weights $\lambda_\mathrm{bald}^\mathrm{rec}$ and $\lambda_\mathrm{bald}^\mathrm{sds}$ balance their contributions.
We define the face region mask as $\hat{\mat{M}}_\mathrm{face} = (1 - \hat{\mat{M}}_\mathrm{hair}) \odot \mat{M}_\mathrm{face}$, where $\hat{\mat{M}}_\mathrm{hair}$ is a dilated hair mask that excludes areas near the hairline, removing shadows and ensuring a smooth transition between the face and bald texture.
To build a diffusion prior from dome-captured human images~\cite{martinez2024ava}, we train an image-to-image inpainting latent diffusion model~\cite{rombach2022ldm} with ControlNet~\cite{zhang2023controlnet}. The diffusion model takes an image prompt, while ControlNet receives a masked input to guide inpainting. For each capture view, we generate a bald image prompt $\mat{I}_\mathrm{cond}$ using a pretrained text-to-image inpainting model~\cite{rombach2022ldm} and apply SDS loss:
\begin{equation}
\label{eq:bald_sds}
\begin{aligned}
    &\nabla_{\theta} \mathcal{L}_\mathrm{bald}^\mathrm{sds}(\vec{z}_t, \mat{I}_\mathrm{cond}, \mat{I}_\mathrm{masked}, t) \\
    &= \mathop{\mathbb{E}} 
    \Big[
        \omega(t) \big( \hat{\epsilon_{\phi}}(\vec{z}_t; \mat{I}_\mathrm{cond}, \mat{I}_\mathrm{masked}, t) - \epsilon \big)
        \frac{\partial \vec{z}_t}{\partial \vec{x}} 
        \frac{\partial \vec{x}}{\partial \theta}
    \Big],
\end{aligned}
\end{equation}
where $\mat{I}_\mathrm{cond}$ is the bald image prompt, $\mat{I}_\mathrm{masked}$ is the hair-masked input image for ControlNet, $t$ is a diffusion timestep, and $\omega(t)$ is the diffusion scheduler weight~\cite{ho2020ddpm}.

\paragraph{Bald image composition}
Optimizing a bald texture MLP for every frame is impractical. Instead, we optimize the MLP for a single neutral expression frame and blend the original image with the rendered bald mesh to generate synthetic bald images across expressions. 
Following HairMapper~\cite{wu2022hairmapper}, we dilate and blur the hair mask to remove shadows and smooth transitions between the haired image $\mat{I}_{\mathrm{orig}}$ and the rendered bald mesh $\mesh{M}_{\mathrm{bald}}$, yielding the processed mask $\tilde{\mat{M}}_{\mathrm{hair}}$. The final synthetic bald image is obtained as:
\begin{equation}
\label{eq:bald_comp}
    \mat{I}_{\mathrm{bald}}=
    \tilde{\mat{M}}_{\mathrm{hair}} \odot 
    \mat{I}_{\mathrm{orig}} + 
    (1 - \tilde{\mat{M}}_{\mathrm{hair}}) \odot \mathcal{R}\left(\mesh{M}_{\mathrm{bald}}\right),
\end{equation}
where $\mathcal{R}$ is a mesh renderer~\cite{pidhorskyi2025drtk}.
  
\subsection{Training}
\label{subsection:training}
We train our compositional universal avatar model using multi-view video captures of people with known point light patterns, following URAvatar~\cite{li2024uravatar}. We modify the loss function from prior work~\cite{saito2024relightable,li2024uravatar} as follows:
\begin{equation} 
    \mathcal{L} = 
    \mathcal{L}_{\mathrm{rec}} + \mathcal{L}_{\mathrm{seg}} + \mathcal{L}_{\mathrm{reg}} +  \mathcal{L}_{\mathrm{kl}}.
\end{equation}
$\mathcal{L}_{\mathrm{rec}}$ ensures image and bald geometry reconstruction, $\mathcal{L}_{\mathrm{seg}}$ promotes separation of face and hair Gaussians, $\mathcal{L}_{\mathrm{reg}}$ constrains Gaussian attributes, and $\mathcal{L}_{\mathrm{kl}}$ applies KL-divergence loss~\cite{kingma2014vae} on expression and hair motion codes from \cref{eq:sample_latent}.

\paragraph{Reconstruction loss.}
To build compositional avatars with separate face and hair Gaussians, we render the avatars in two ways: face-only rendering, which uses only face and eye Gaussians, and compositional rendering, which includes all Gaussians. L1 and SSIM losses are applied to both renderings. Face-only rendering compares the rendered output $\hat{\mat{I}}_{\mathrm{face}}$  with the synthetic bald images $\mat{I}_{\mathrm{bald}}$ (\cref{eq:bald_comp}), ensuring that face Gaussians reconstruct the full facial appearance, including occluded scalp regions. Compositional rendering compares the rendered output $\hat{\mat{I}}_{\mathrm{comp}}$ with the studio-captured images $\mat{I}_{\mathrm{orig}}$, ensuring that face and hair Gaussians together reconstruct the full avatar appearance with a natural transition between components.
For bald geometry reconstruction, we apply an L2 loss between the predicted bald mesh vertices $\hat{\mat{V}}_{\mathrm{b}}$ (\cref{eq:dec_bald}) and the groundtruth bald tracked mesh vertices $\mat{V}_{\mathrm{b}}$.

\paragraph{Boundary-free segmentation loss.}
The rendering losses alone are insufficient to disentangle face and hair Gaussians. We incorporate a Gaussian segmentation loss inspired by GALA~\cite{kim2024gala}, assigning one-hot labels $[1, 0]$ and $[0, 1]$ to hair and face Gaussians and splatting them into a segmentation map $\hat{\mat{S}}\in \mathbb{R}^{H \times W \times 2}$. The loss is computed as the L1 loss between $\hat{\mat{S}}$ and the groundtruth segmentation mask $\mat{S}$.  
To avoid sharp segmentation boundaries, we stop applying the segmentation loss near the face-hair boundary after a certain number of iterations. This excluded region is determined by expanding both the hair and face masks and identifying the overlap. This allows a natural blending between face and hair regions, enabling a seamless composition.

\subsection{Applications}
\label{subsection:application}
Our compositional universal avatar model serves as a unified model for training subjects, similar to URAvatar~\cite{li2024uravatar}. Beyond unification, its compositional nature enables flexible hairstyle swapping, functioning as a Gaussian-based 3D hair salon~\cite{hu2015hairsalon}, where hairstyles can seamlessly transfer between avatars.
Additionally, our model serves as a powerful prior for few-shot fine-tuning, enabling hair-compositional avatars for novel identities. It maintains independent control over face and hair, supporting bidirectional hairstyle transfer from the training subjects to the novel identity and vice versa, offering greater flexibility in avatar customization.

\paragraph{Hairstyle transfer.}
Trained on a large corpus of multi-view human images, our model enables compositional avatar generation with independently controllable face and hair.
During training, the same identity is used for both face and hair hypernetworks, as subjects are captured with their own hairstyle. At test time, face and hair identity data can differ, allowing flexible composition.
Hair Gaussians are defined relative to the head and adapt automatically to the target's shape via the bald mesh anchor (\cref{eq:add_trans}), enabling seamless transfer without scaling or alignment~\cite{wang2025mega}.

\paragraph{Few-shot personalization.}
Our model serves as a strong prior for creating personalized avatars with few-shot finetuning. Unlike pretraining, we bypass the need for synthetic bald images (\cref{subsection:baldify}) by applying the face-only rendering loss only to visible regions, as the prior keeps face Gaussians plausible even under occlusions. The compositional rendering loss remains unchanged, using the full image, while gradients from face Gaussians are detached to prevent them from reconstructing hair regions.
To leverage the prior, we only update the last layer of the hypernetworks.

\section{Experiments}
\label{sec:experiments}

\paragraph{Dataset.} 
We utilize a capture system inspired by prior work on creating relightable 3D avatars~\cite{saito2024relightable,li2024uravatar}. The system captures calibrated and synchronized multi-view images at a resolution of \(4096 \times 2668\), using 110 cameras and 460 white LED lights operating at 90 Hz. Each subject performs a predefined sequence of facial expressions during the capture process. 
To collect diverse illumination patterns, we employ time-multiplexed lighting, where grouped or random sets of 5 lights are activated per frame, interleaved with fully-lit frames every 3 frames to ensure stable facial mesh tracking. Fully-lit frames are processed using topologically consistent coarse mesh tracking~\cite{saito2024relightable}, while partially lit frames use interpolated tracked meshes, head poses, UV texture maps, and gaze data from adjacent fully-lit frames. 
We capture 260 subjects in total, and 252 subjects out of them are used for the pretraining stage, and 8 subjects are used for evaluation of the fine-tuning stage.

\paragraph{Training details.} 
We adopt the architecture of URAvatar~\cite{li2024uravatar} for both the face and hair hypernetworks and decoders. The hypernetworks utilize a UNet-based architecture~\cite{RFB15unet}, where multi-level feature maps are produced and added to the outputs of each layer in the decoder. For the bald geometry predictor, we use a convolutional encoder and an MLP-based decoder to predict the vertex positions.

For training the compositional prior model, we use the Adam optimizer~\cite{kingma2014adam} with a learning rate of $5 \times 10^{-4}$. The training process involves 64 NVIDIA A100 GPUs with a per-GPU batch size of 1 for 250k iterations, taking approximately 7 days. 
During the fine-tuning stage for personalization, the learning rate is set to $10^{-5}$, and we use a single NVIDIA A100 GPU with a batch size of 1 for 2k iterations. To ensure accurate reconstruction, only the face hypernetwork is trained for the first 400 iterations, while the hair hypernetwork is trained for the remaining iterations to prevent hair Gaussians from reconstructing facial regions. The fine-tuning process takes approximately 1 hour.
For the ablation study, we train on 16 subjects to isolate key component effects while keeping computation manageable.

\begin{figure}[t]
    \centering
    \begin{subfigure}[b]{0.24\columnwidth}
         \centering
         \includegraphics[width=\textwidth]{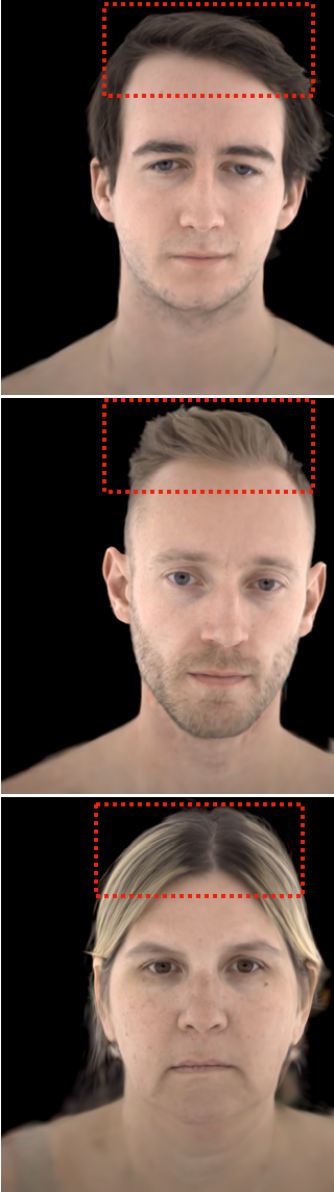}
         \caption{Ours}
         \label{fig:comparison_ours}
    \end{subfigure}
    \begin{subfigure}[b]{0.24\columnwidth}
         \centering
         \includegraphics[width=\textwidth]{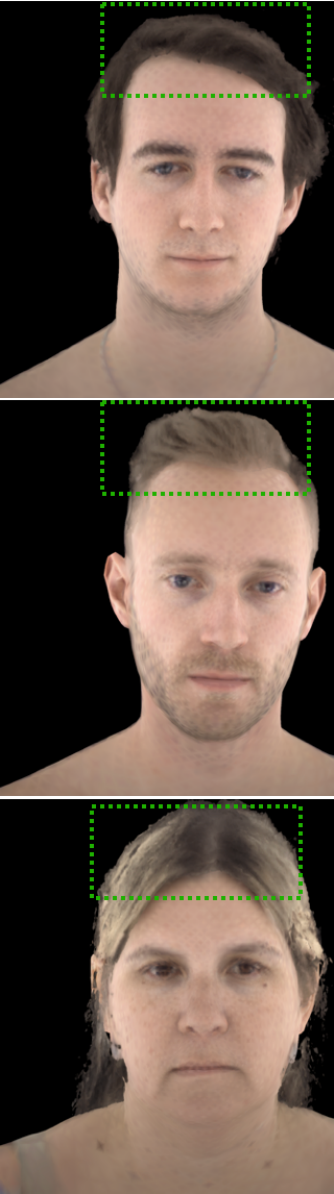}
         \caption{DELTA}
         \label{fig:comparison_delta}
    \end{subfigure}
    \begin{subfigure}[b]{0.24\columnwidth}
         \centering
         \includegraphics[width=\textwidth]{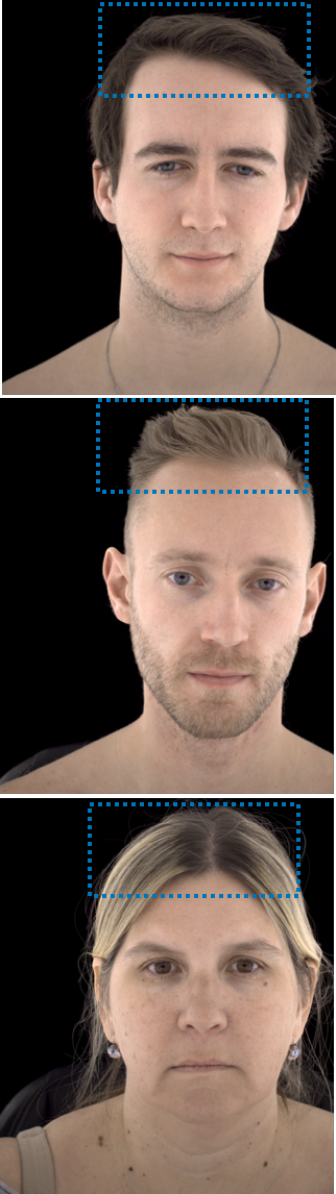}
         \caption{GT}
         \label{fig:comparison_gt}
    \end{subfigure}
    \begin{subfigure}[b]{0.2049\columnwidth}
         \centering
         \includegraphics[width=\textwidth]
         {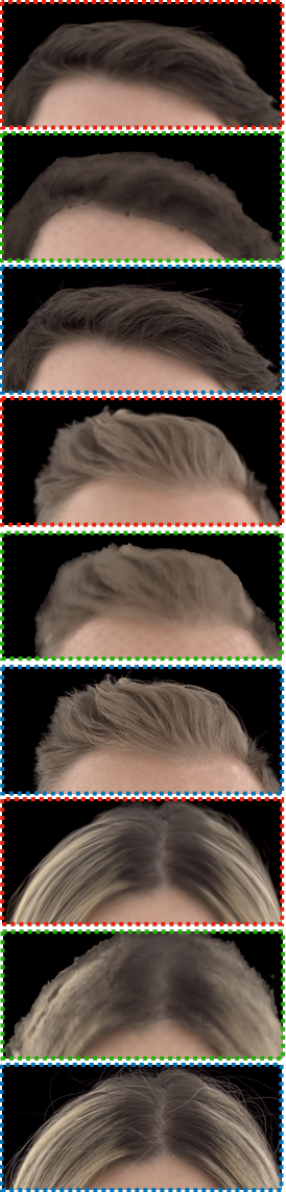}
         \caption{Zoom-in}
         \label{fig:comparison_zoom}
    \end{subfigure}
    \caption{\textbf{Qualitative comparison with DELTA~\cite{Feng2023DELTA}.} The compositional avatar generated by finetuning our universal model with monocular capture of novel identity shows more realistic face and hair details than DELTA. Best viewed on a digital display.}
    \label{fig:comparison_delta}
\end{figure}

\subsection{Comparison}
\label{subsection:comparison}
\paragraph{Compositional avatar personalization.}
We evaluate our model’s effectiveness as a prior for compositional avatar personalization by comparison with DELTA~\cite{Feng2023DELTA}, a monocular video-based method. Using fully-lit, front-view head rotation sequences from 8 unseen identities, we fine-tune our model per subject for a fair comparison. Since DELTA is a single-subject avatar model, we compute average scores across all subjects using L1 loss, PSNR, SSIM, and LPIPS.

\begin{table}[t]

\centering
\small{

\begin{tabular}{lcccc}
\toprule
       & L1 $\downarrow$ 
       & PSNR $\uparrow$ 
       & SSIM $\uparrow$
       & LPIPS $\downarrow$ \\
\midrule
DELTA~\cite{Feng2023DELTA} & $0.0344$ & $23.775$ & $0.790$ & $0.0241$ \\
Ours & $\mathbf{0.0223}$ & $\mathbf{27.040}$ & $\mathbf{0.817}$ & $\mathbf{0.0131}$ \\
\bottomrule
\end{tabular}
}
\caption{\textbf{Quantitative comparison with DELTA~\cite{Feng2023DELTA}.} }

\label{tab:quant_delta}
\end{table}

\Cref{tab:quant_delta} presents quantitative results demonstrating that avatars generated by fine-tuning our model achieve superior performance across all metrics compared to DELTA. Specifically, our method surpasses DELTA in both facial and hair details, as shown qualitatively in \cref{fig:comparison_delta}. DELTA exhibits limitations in facial quality due to the restricted expressiveness of its parametric face model, showing blocky artifacts. Furthermore, its reliance on an LBS-based deformation model for hair degrades the fine details.

\begin{figure}[t]
\includegraphics[width=0.98\linewidth, trim={0cm 0cm 1cm 1cm}]{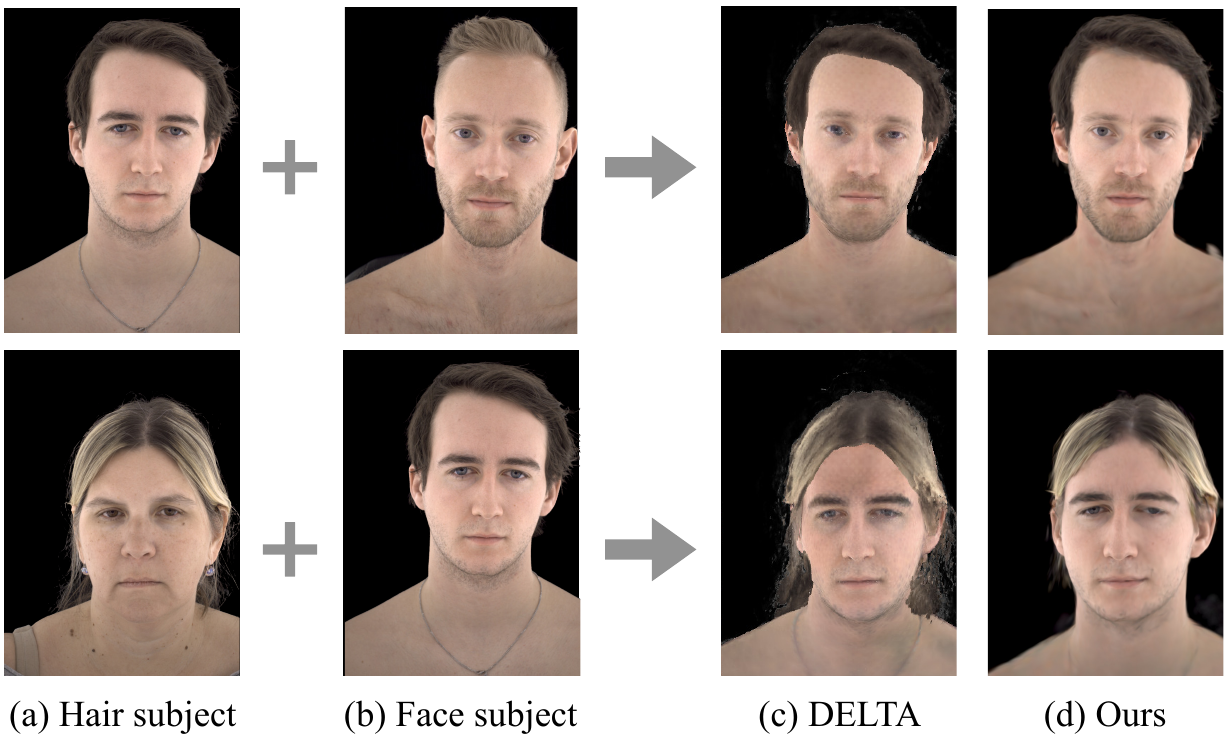}
\vspace{-1mm}
\caption{\textbf{Qualitative comparison of hairstyle transfer.} Our method achieves more natural hairstyle transfer with seamless face-hair blending, while DELTA creates a rigid boundary. Additionally, our results offer greater visual fidelity in face and hair.}
\label{fig:comparison_swap}
\end{figure}

In addition, \cref{fig:comparison_swap} highlights the hair-swapping results between different avatars. Our method achieves this by combining the face hypernetwork of one finetuned avatar with the hair hypernetwork of another, whereas DELTA performs swapping by pairing the face mesh of one avatar with the hair NeRF of another. However, DELTA's reliance on segmentation masks enforces rigid face-hair boundaries, resulting in unnatural transitions at the boundary. In contrast, our approach utilizes a boundary-free segmentation loss, enabling smooth and natural hair transfer between avatars.

\paragraph{Interpolation.}
\begin{figure}[t]
\includegraphics[width=0.95\linewidth, trim={0cm 0cm 1cm 1cm}]{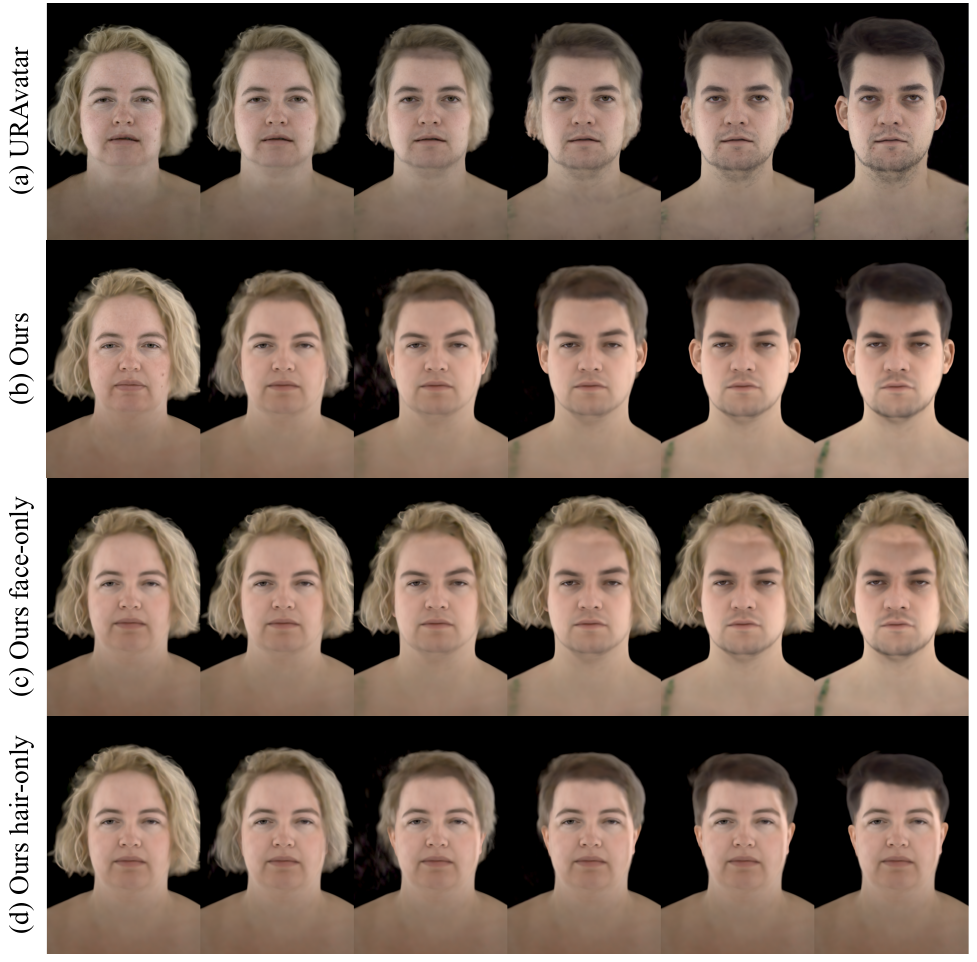}
\caption{\textbf{Comparison of avatar interpolation.} URAvatar, which models the face and hair holistically, exhibits unnatural transitions from hair to ear during interpolation, leading to broken ears (a). In contrast, HairCUP achieves more natural interpolation by disentangling face and hair modeling (b). (c) and (d) present separate interpolation results of HairCUP for the face and hair, respectively.}
\label{fig:interpolation}
\end{figure}

To evaluate compositional modeling in learning the manifold of 3D head avatars, we compare interpolation results with URAvatar, a holistic 3D avatar model. As shown in \cref{fig:interpolation}, we interpolate between two avatars by blending bias maps from hypernetworks and the anchor geometry.
URAvatar struggles with interpolation between subjects with long and short hair, causing ears to morph into hair and distort facial structure. In contrast, our method yields smooth transitions in both face and hair, as evident in face-only interpolation. This demonstrates effective disentanglement of face and hair manifolds, enabling better semantic correspondence and smoother transitions across subjects, while holistic modeling suffers from entangled representations that obscure facial structure.

\subsection{Ablation Study}
\label{sec:ablation}

\begin{figure}[t]
  \centering
   \includegraphics[width=0.9\linewidth]{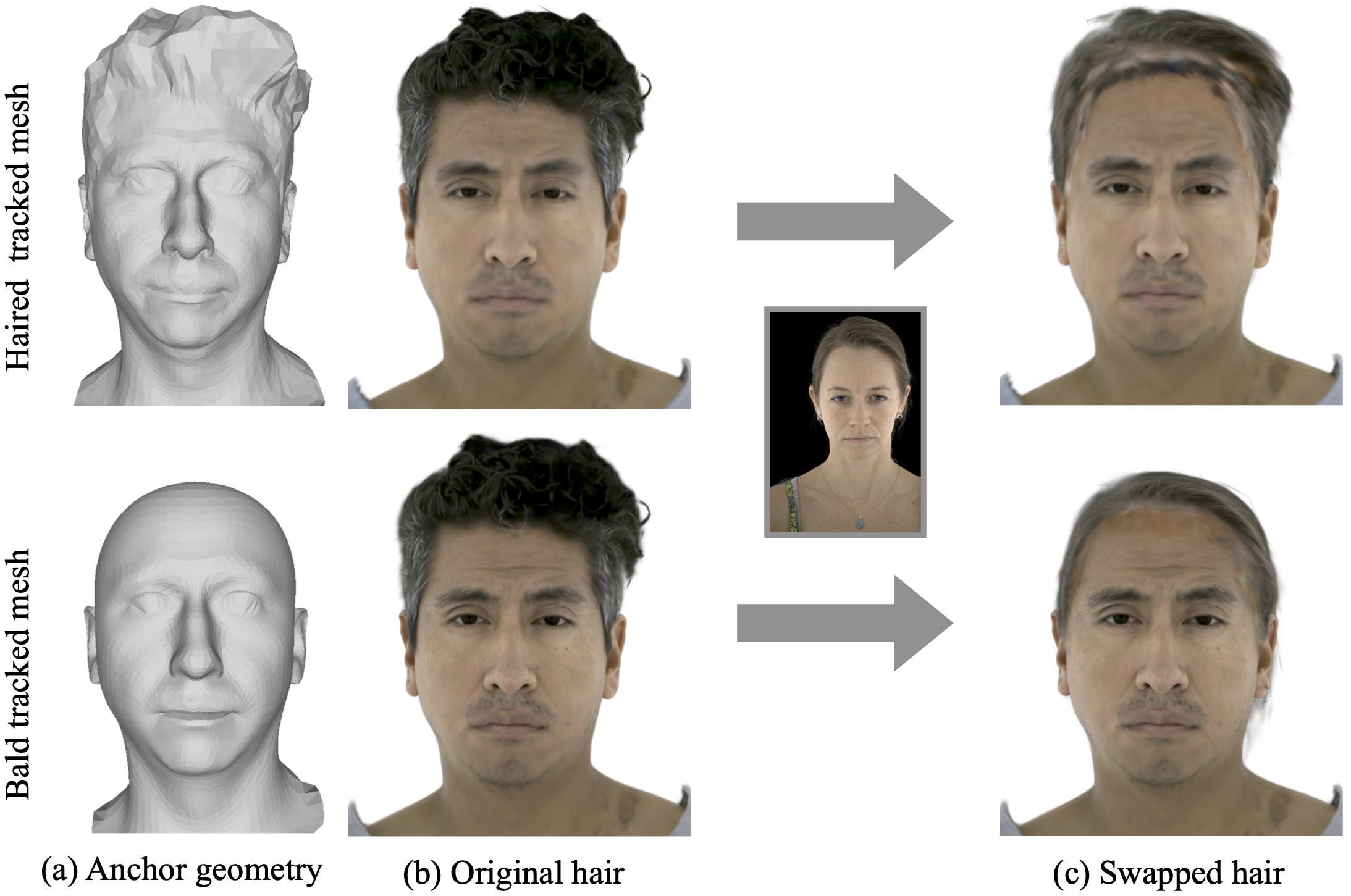}
   \caption{\textbf{Ablation on anchor geometry.} Anchoring 3D Gaussians to a haired mesh leads to unnatural transfers due to misaligned source geometry. Using a bald mesh enables more natural hairstyle transfer by providing a consistent scalp anchor.}
   \label{fig:ablation_anchor}
\end{figure}

\paragraph{Effect of hair offset modeling.}
We anchor hair Gaussians on the UV geometry map of the neutral bald mesh (\cref{eq:add_trans}). \Cref{fig:ablation_anchor} compares hair-swapping results using a haired tracked mesh anchor versus a bald tracked mesh anchor. While the haired tracked mesh anchor, used in prior works~\cite{saito2024relightable,li2024uravatar,qian2024gaussianavatars}, performs well for original hair, it often misaligns when swapping hair, causing unnatural transfers.
In contrast, anchoring Gaussians to the bald tracked mesh ensures hair placement relative to the scalp, maintaining compatibility with the target avatar’s head. This enables smooth and natural hair transfer, adapting seamlessly to diverse head shapes and sizes (\cref{fig:teaser}).

\begin{figure}[t]
  \centering
   \includegraphics[width=0.95\linewidth]{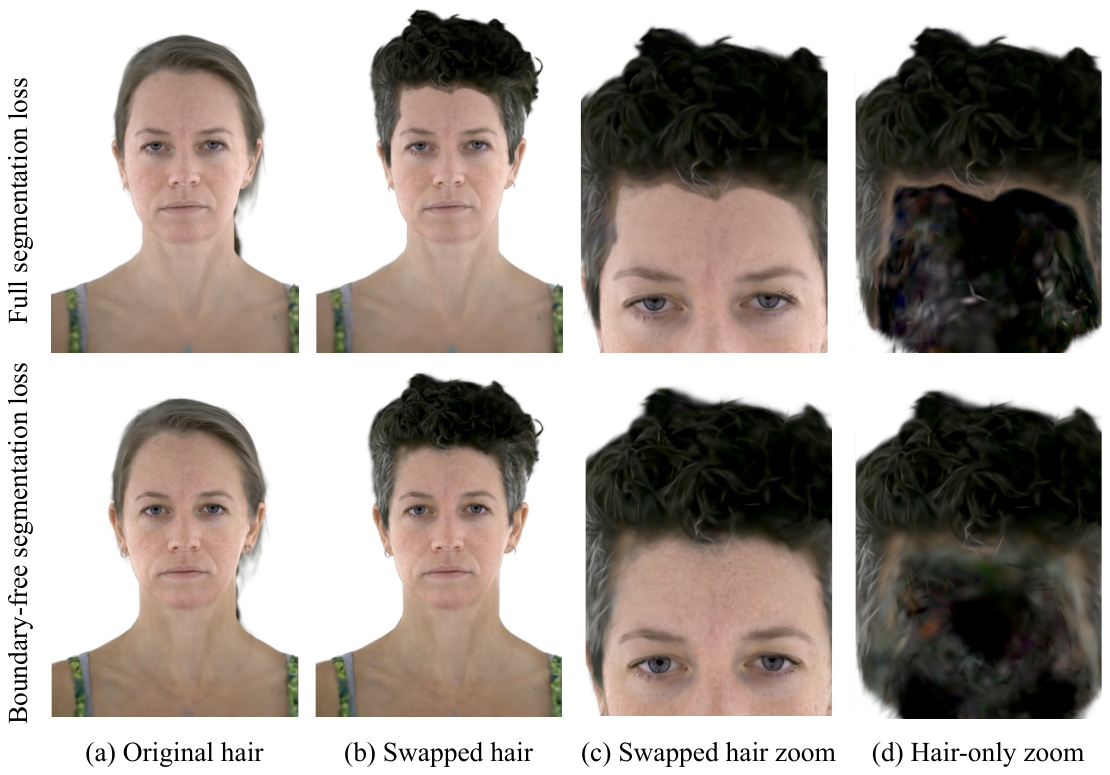}
   \caption{\textbf{Ablation study on boundary-free segmentation loss.} 
   Boundary-free segmentation loss yields smoother face-hair transitions, while full-region loss creates sharp boundaries.
   }
   \label{fig:ablation_segm}
\end{figure}

\paragraph{Segmentation loss.}
To evaluate the effect of boundary-free segmentation loss, we compare two settings in \cref{fig:ablation_segm}: applying segmentation loss throughout training versus disabling it near the face-hair boundary midway. Defining this boundary is inherently difficult, as the transition between face and hair is often gradual, whereas segmentation masks enforce hard-edged supervision. As a result, applying the loss across the entire region can produce unnaturally sharp face-hair boundaries. In contrast, relaxing the loss near the boundary leads to smoother transitions and more natural, seamless hairstyle transfer.

\section{Conclusion}
\label{sec:conclusion}

We introduced a compositional universal avatar model that independently represents face and hair, addressing limitations of holistic approaches. By leveraging synthetic bald image generation, the model learns disentangled latent spaces, enabling realistic hair transfer and few-shot avatar personalization. Experiments demonstrate effective separation and transfer of face and hair with high fidelity,  supporting flexible and realistic avatar customization. 

\noindent\textbf{Limitations.} 
Our method doesn’t account for large hair dynamics, as the training data mostly contains stable motion.
Synthetic bald images may introduce color discrepancies, as the bald texture is optimized from a single fully-lit frame.
Additionally, Gaussians anchored on a fixed UV map can lead to quality variations across different hairstyles.
Finally, the occluded scalp remains hard to relight due to limited supervision under partial lighting.

\section*{Acknowledgments}
The work of Kim and Joo is supported in part by NRF grant funded by the Korean government (MSIT) (No. RS-2022-NR070498), and IITP grant funded by the Korea government (MSIT) [No. RS-2024-00439854 and RS-2021-II211343].
{
    \small
    \bibliographystyle{ieeenat_fullname}
    \bibliography{haircup_ref}
}

\ifarxiv
    \maketitlesupplementary

\appendix

\section{Preliminaries: URAvatar~\cite{li2024uravatar}}
Our method builds on URAvatar~\cite{li2024uravatar}, a universal 3D avatar model that extends the relightable 3D Gaussian representation of \citet{saito2024relightable} to multiple subjects. To provide the necessary background, we first describe \citeauthor{saito2024relightable}'s method for person-specific relightable 3D Gaussian avatars, followed by the multi-subject extension introduced in URAvatar~\cite{li2024uravatar}.

\citet{saito2024relightable} proposed a relightable 3D Gaussian head avatar model~\cite{saito2024relightable} that learns a latent space of facial expressions using a conditional variational autoencoder (VAE)~\cite{kingma2014vae}. The encoder maps an unwrapped UV texture map and tracked mesh vertices to an expression code, which is then used by a set of decoders to generate 3D Gaussian primitives.
Given the unwrapped texture map $\mat{T}$ and tracked mesh vertices $\mat{V}$, the encoder produces the mean $\mu_e$ and covariance $\sigma_e$ of the expression code:
\begin{equation}
    \label{eq:encoder}
    \mu_{e}, \sigma_{e}=\mathcal{E}(\mat{V}, \mat{T}; \Theta_{e}).
\end{equation}
The decoders reconstruct the tracked mesh vertices $\mat{V}$ and generate Gaussian primitives~\cite{kerbl20233dgs}, which are splatted~\cite{zwicker2002ewa} to render the avatar.
Building on this, URAvatar~\cite{li2024uravatar} generalizes the relightable 3D Gaussian avatar to multiple subjects by introducing an identity-conditioned hypernetwork~\cite{cao2022authentic}. This hypernetwork, $\mathcal{E}_{\mathrm{id}}$, generates bias maps for avatar decoders and expression-agnostic attributes, given the UV-unwrapped mean albedo and geometry maps of the facial tracked meshes:
\begin{equation}
\label{eq:hyper}
    \Theta^{\mathrm{id}}_{\mathrm{g}},
    \Theta^{\mathrm{id}}_{\mathrm{vi}},
    \Theta^{\mathrm{id}}_{\mathrm{vd}},
    \left\{o_{k}, \vec{\rho}_{k}\right\}^{N}_{k=1}=
    \mathcal{E}_{\mathrm{id}}(\mat{T}_\mathrm{mean}, \mat{G}_\mathrm{mean};\Phi_{\mathrm{id}}),
\end{equation}
where $N$ is the number of Gaussians, $o_{k}$ and $\vec{\rho}_{k}$ are expression-agnostic opacity and albedo of 3D Gaussians, and $\Theta^{\mathrm{id}}_{\mathrm{g}},
\Theta^{\mathrm{id}}_{\mathrm{vi}},\Theta^{\mathrm{id}}_{\mathrm{vd}}$ are identity-conditioned bias maps injected into the intermediate feature maps of their respective decoders.

The geometry decoder $\mathcal{D}_{\mathrm{g}}$ predicts tracked mesh vertices:
\begin{equation}
\label{eq:dec_geo}
    \left\{\hat{t}_{k}\right\}_{k=1}^{N}=
    \mathcal{D}_{\mathrm{g}}(\vec{z}, \vec{e}_{\mathrm{\{l, r\}}},\vec{r}_{n};\Theta_{\mathrm{g}}^{\mathrm{id}},\Phi_{\mathrm{g}}),
\end{equation}
where $\vec{z}$ is an expression code, $\vec{e}_{\mathrm{\{l,r\}}}$ are eye gaze direction vectors, and $\vec{r}_{n}$ is the axis-angle neck rotation relative to the head. The predicted vertices serve as anchors for Gaussians produced by the appearance decoders.
The two Gaussian decoders, $\mathcal{D}_\mathrm{vi}$ and $\mathcal{D}_\mathrm{vd}$, generate the geometric and appearance attributes needed to evaluate each Gaussian’s radiance:
\begin{equation}
\label{eq:dec_gs_vi}
    \{\delta{\vec{t}}_{k}, \vec{q}_{k}, \vec{s}_{k}, \vec{d}_{k},
    \sigma_{k}\}_{k=1}^{N} =
    \mathcal{D}_{\mathrm{vi}}(
    \vec{z},\vec{e}_{\mathrm{ \{ l,r \}}}, \vec{r}_{n};\Theta_{\mathrm{vi}}^{\mathrm{id}}, \Phi_{\mathrm{vi}}),
\end{equation}
\begin{equation}
\label{eq:dec_gs_vd}
    \left\{
    \delta{\vec{n}}_{k},
    v_{k}\right\}_{k=1}^{N} =
    \mathcal{D}_{\mathrm{vd}}(
    \vec{z},\vec{e}_{\mathrm{ \{ l,r \} }}, \vec{r}_{n}, \vec{\omega}_{o};
    \Theta_{\mathrm{vd}}^{\mathrm{id}}, \Phi_{\mathrm{vd}}),
\end{equation}
where $\delta\vec{t}_{k}$ is the position offset, $\vec{q}_{k}$ is the orientation, and $\vec{s}_{k}$ is the scale of each Gaussian. $\vec{d}_{k}$ represents the SH coefficients for color and monochrome components~\cite{saito2024relightable}, and $\sigma_{k}$ is the roughness parameter as defined in \cref{eq:specular} of the main paper. The term $\delta\vec{n}_{k}$ denotes the view-dependent delta normal, and $v_{k}$ represents the visibility term.

To account for eye modeling, URAvatar includes a universal relightable explicit eye model adapted from \citet{saito2024relightable}. The eye hypernetwork $\mathcal{E}_{\mathrm{eye}}$ generates bias maps for the eye Gaussian decoders, ensuring identity-specific adaptation:
\begin{equation}
\label{eq:eye_hyper}
    \Theta_{\mathrm{vi}}^{\mathrm{e}},
    \Theta_{\mathrm{vd}}^{\mathrm{e}}=
    \mathcal{E}_{\mathrm{eye}}
    (
        \mat{T}_{\mathrm{e}},
        \mat{G}_{\mathrm{e}};
        \Phi_{\mathrm{id}}^{\mathrm{e}}
    ),
\end{equation}
where $\mat{T}_{\mathrm{e}}$ and $\mat{G}_{\mathrm{e}}$ correspond to the eye region in the mean texture and geometry maps. The eye Gaussian decoders predict similar attributes as the main avatar decoders, with a unified decoder for the specular visibility map to better preserve eye reflection priors.
For further details, we refer readers to the paper~\cite{li2024uravatar}.

\section{Synthetic Bald Image Generation}
\subsection{Synthetic Bald Image Pairs}
To validate the consistency between the original and synthetic bald images used for training, we present example image pairs in \cref{fig:bald_pair_supp}. These pairs are constructed using the compositing scheme described in the main paper (\cref{eq:bald_comp}), where the face region is taken from the original image and the occluded scalp region is inpainted with the rendered bald mesh. By carefully processing the hair mask to ensure smooth transitions, our method produces visually coherent bald images across diverse viewpoints and expressions.

\begin{figure*}[t]
  \centering
  \includegraphics[width=\linewidth]{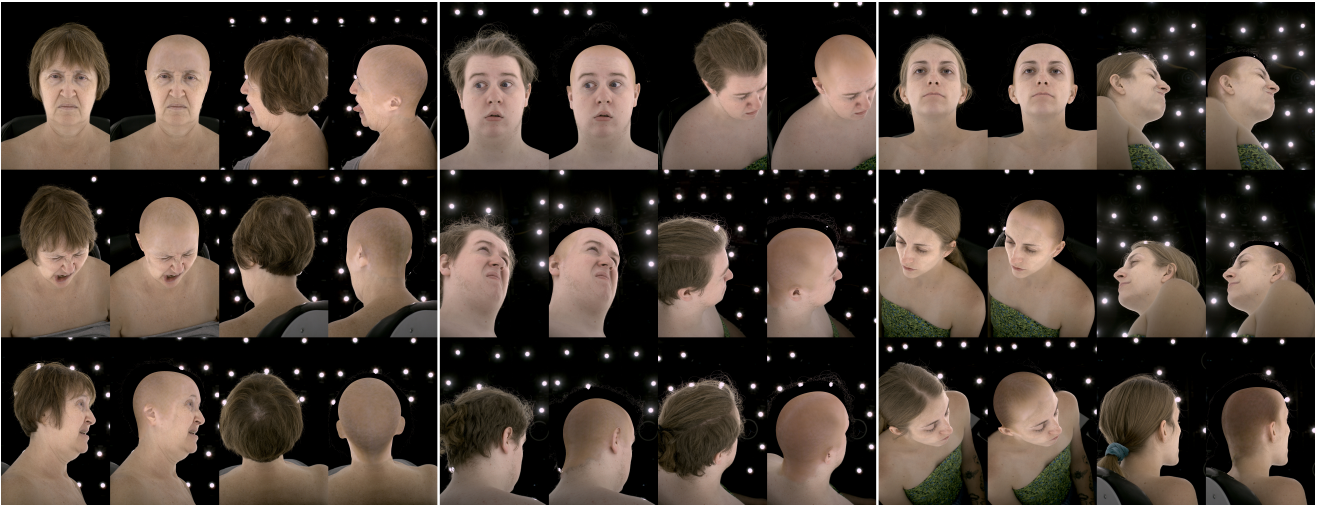}
  \caption{\textbf{Synthetic bald image pairs.} Each pair shows (left) the original image and (right) the synthetic bald image generated using our compositing pipeline. The synthetic bald images preserve facial identity while removing hair occlusion, enabling effective supervision for face-hair disentanglement.}
  \label{fig:bald_pair_supp}
\end{figure*}

\subsection{Bald Texture Optimization}
\paragraph{Optimization details.} We present the details of bald texture optimization (\cref{subsection:baldify}). To optimize the bald texture MLP, we use the loss function $\mathcal{L}_{\mathrm{bald}}$ from \cref{eq:bald_optim} in the main paper, running a two-stage optimization over 2500 iterations.
For the first 1500 iterations, we apply only the reconstruction loss $\mathcal{L}_{\mathrm{bald}}^{\mathrm{rec}}$ to reconstruct the visible face region. In the next 1000 iterations, we introduce SDS loss~\cite{poole2023dreamfusion} to refine the texture in hair-occluded regions while continuing reconstruction loss, using weights $\lambda_{\mathrm{bald}}^{\mathrm{rec}}=1$ and $\lambda_{\mathrm{bald}}^{\mathrm{sds}}=10^{-6}$.
During the SDS stage, we employ an inpainting image-to-image diffusion model with ControlNet~\cite{zhang2023controlnet}, trained on dome-captured human images~\cite{martinez2024ava}.
For the first 500 iterations of SDS loss, we use a bald image prompt generated from a pretrained text-to-image (T2I) inpainting diffusion model~\cite{rombach2022ldm} as an input image prompt to our diffusion model. In the final 500 iterations, we replace this image prompt with the rendered bald mesh, using its actively optimized texture map for rendering. By this stage, the rendered bald image provides better consistency than the bald image generated from the pretrained T2I model, leading to more coherent texture refinement.

\begin{figure}
     \centering
     \begin{subfigure}[b]{0.27342\columnwidth}
         \centering
         \includegraphics[width=\textwidth]{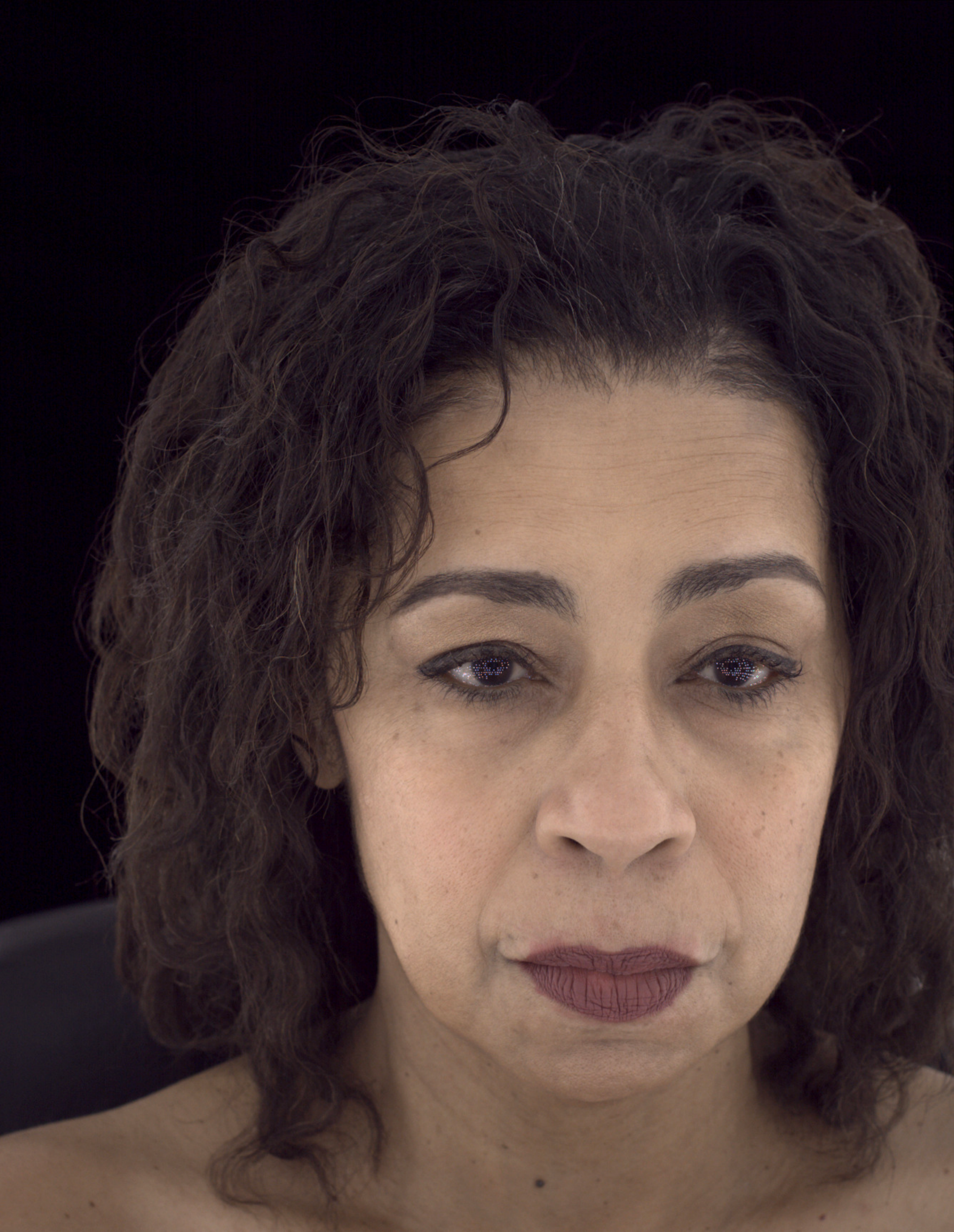}
         \caption{Target hairstyle}
         \label{fig:front}
     \end{subfigure}
     \begin{subfigure}[b]{0.27342\columnwidth}
         \centering
         \includegraphics[width=\textwidth]{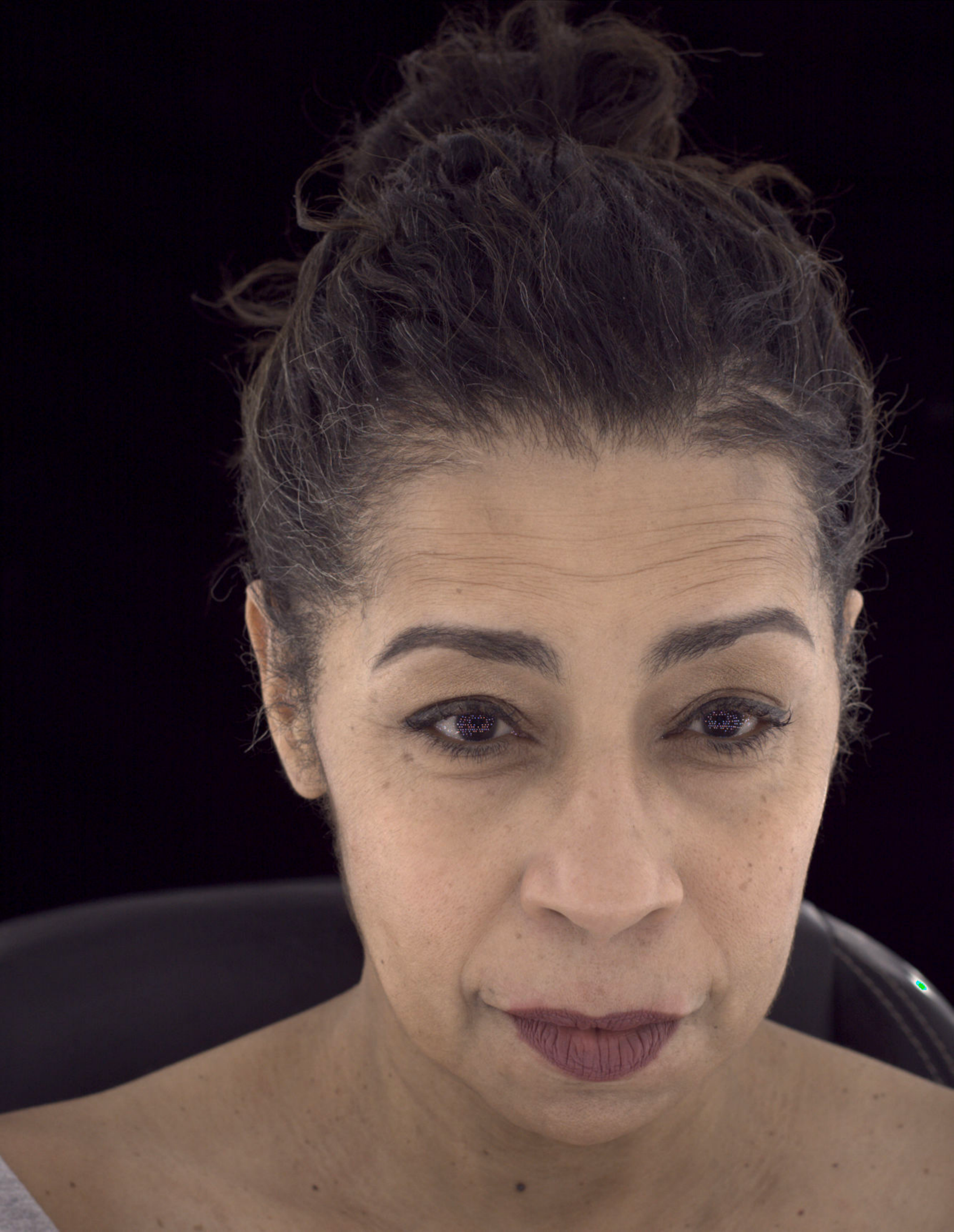}
         \caption{Hair-tied capture}
         \label{fig:front_ear}
     \end{subfigure}
     \begin{subfigure}[b]{0.353\columnwidth}
         \centering
         \includegraphics[width=\textwidth]{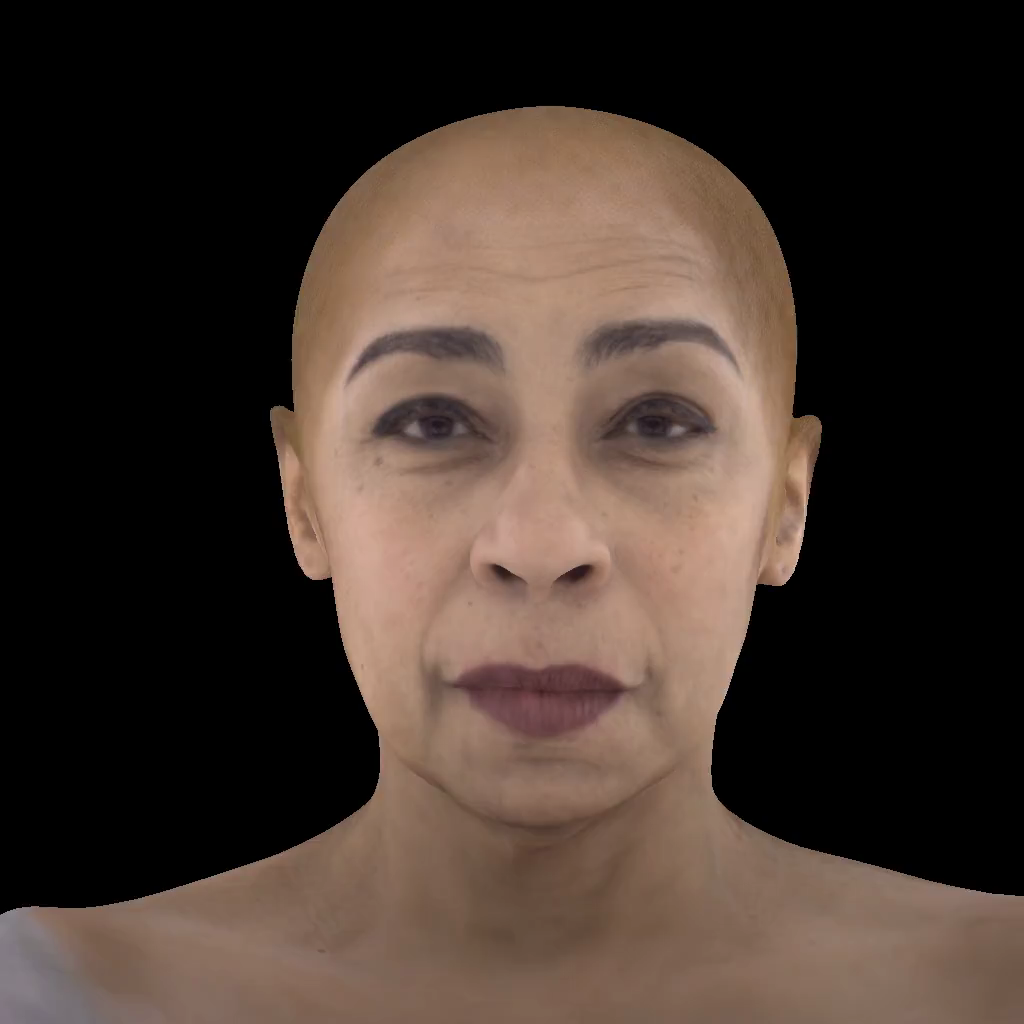}
         \caption{Optimized bald mesh}
         \label{fig:bald_mesh}
     \end{subfigure}
    \caption{\textbf{Auxiliary capture for bald texture optimization.} To minimize occlusion from certain hairstyles, we capture an additional image with the subject’s hair tied back (b). This ensures that the optimized bald texture (c) maintains consistent skin color, even when the target hairstyle (a) differs.} 
\label{fig:bald_tex_optim}
\end{figure}
\paragraph{Auxiliary capture for bald texture optimization.} 
Certain hairstyles, such as long hair or bangs, can cause severe occlusions that degrade the quality of the optimized texture map. To mitigate this, we capture subjects with their hair tied back or secured with a thin hairband to minimize occlusion (\eg \cref{fig:front_ear}). Reducing occlusion maximizes the visible reconstruction region and decreases reliance on the diffusion prior.
It is important to note that this capture is used solely for bald texture optimization. For instance, in \cref{fig:bald_tex_optim}, although the target hairstyle for training the hair-compositional avatar corresponds to \cref{fig:front}, we use a separate capture (\cref{fig:front_ear}) to ensure that the pseudo-bald images maintain consistent skin color beneath the hair.

\section{More Qualitative Results}
\paragraph{Hairstyle transfer.}
To demonstrate the robust disentanglement and flexible compositionality of our 3D avatar model, we provide additional qualitative results of hairstyle transfer in this section. As elaborated in the main paper, our framework enables the independent manipulation and transfer of facial and hair components across different identities. This is achieved by defining hair Gaussians relative to a bald mesh anchor, which allows for seamless adaptation to the target subject's head shape without the need for additional scaling or alignment steps.
Figure \ref{fig:single_face_multi_hair} illustrates the capability of our model to transfer various hairstyles onto a single facial identity while preserving the facial characteristics and expression. In this example, a consistent facial identity and expression is combined with different hair attributes sourced from various individuals, showcasing how a single avatar can adopt diverse hairstyles realistically. Conversely, Figure \ref{fig:single_hair_multi_face} demonstrates the flexibility of our method in transferring a single hairstyle onto multiple distinct facial identities, each maintaining their unique facial features and expressions. This cross-reenactment with hair transfer highlights the generalizability of our hair model, as it adapts a specific hairstyle to different head shapes and facial features, producing visually coherent and high-fidelity results. These examples collectively emphasize the effectiveness of our compositional prior model in achieving high-quality, controllable 3D avatar synthesis through disentangled face and hair representations.

\begin{figure}[t]
\centering
\includegraphics[width=\columnwidth]{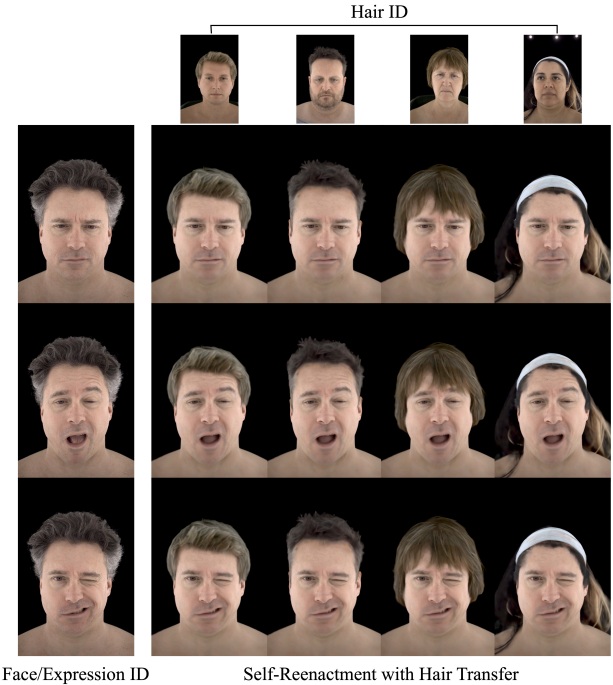}
\caption{\textbf{Hairstyle Transfer: Single Face, Multiple Hairs.} This figure demonstrates transferring various hairstyles onto a single facial identity. The consistent facial features and expressions highlight the model's ability to seamlessly integrate different hairstyles while preserving identity.}
\label{fig:single_face_multi_hair}
\end{figure}

\begin{figure}[t]
\centering
\includegraphics[width=\columnwidth]{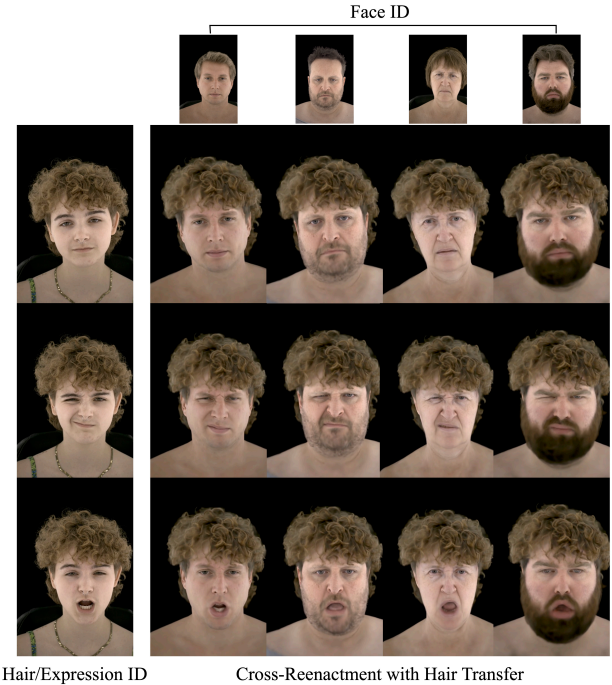}
\caption{\textbf{Hairstyle Transfer: Single Hair, Multiple Faces.} This figure illustrates transferring a consistent hairstyle across multiple distinct facial identities with varying expressions. The results show the adaptability of our hair model to different head shapes, enabling robust cross-reenactment with hair transfer.}
\label{fig:single_hair_multi_face}
\end{figure}

\begin{figure}[t]
    \centering
    \includegraphics[width=\columnwidth]{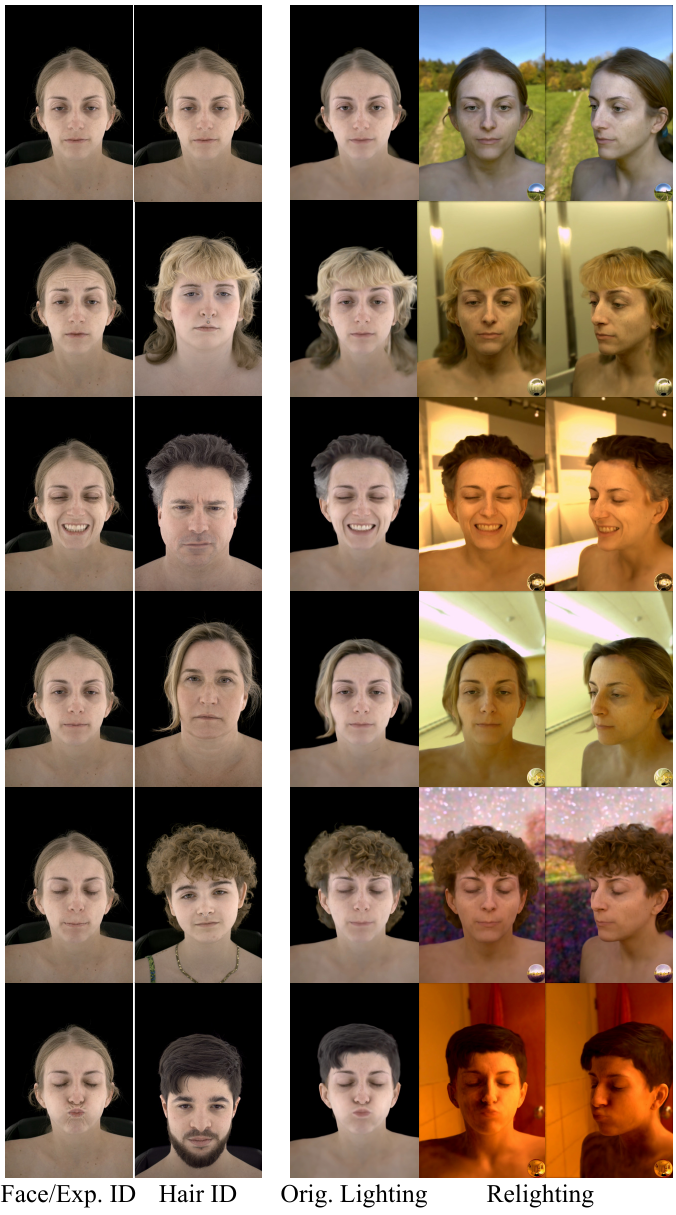}
    \caption{\textbf{Relighting with hairstyle transfer.} The leftmost column shows face and expression reference images captured from a real subject (Face/Exp. ID), with expression changing across frames. The second column shows the hair identity image (Hair ID) used for hair transfer.
The remaining columns visualize avatar rendering results under different lighting conditions.
``Orig. Lighting'' corresponds to the original lighting condition under which the subject was captured. ``Relighting'' corresponds to avatar rendering under novel lighting conditions defined by various environment maps, with each environment map visualized as a reference ball in the bottom-right corner of each image. }
    \label{fig:relighting}
\end{figure}

\paragraph{Relighting with hairstyle transfer.}
Our model inherits the relightable 3D Gaussian appearance model from \citet{saito2024relightable} and URAvatar~\cite{li2024uravatar}, enabling realistic lighting effects on both face and hair. As shown in \cref{fig:relighting}, our approach is the first to support relightable hairstyle transfer, maintaining consistent illumination across both components. While this aspect builds on existing techniques, it marks a significant step forward by demonstrating relightability in the context of hairstyle transfer, ensuring natural and cohesive lighting under varying conditions.

\begin{figure*}[t]
  \centering
   \includegraphics[width=0.9\linewidth]{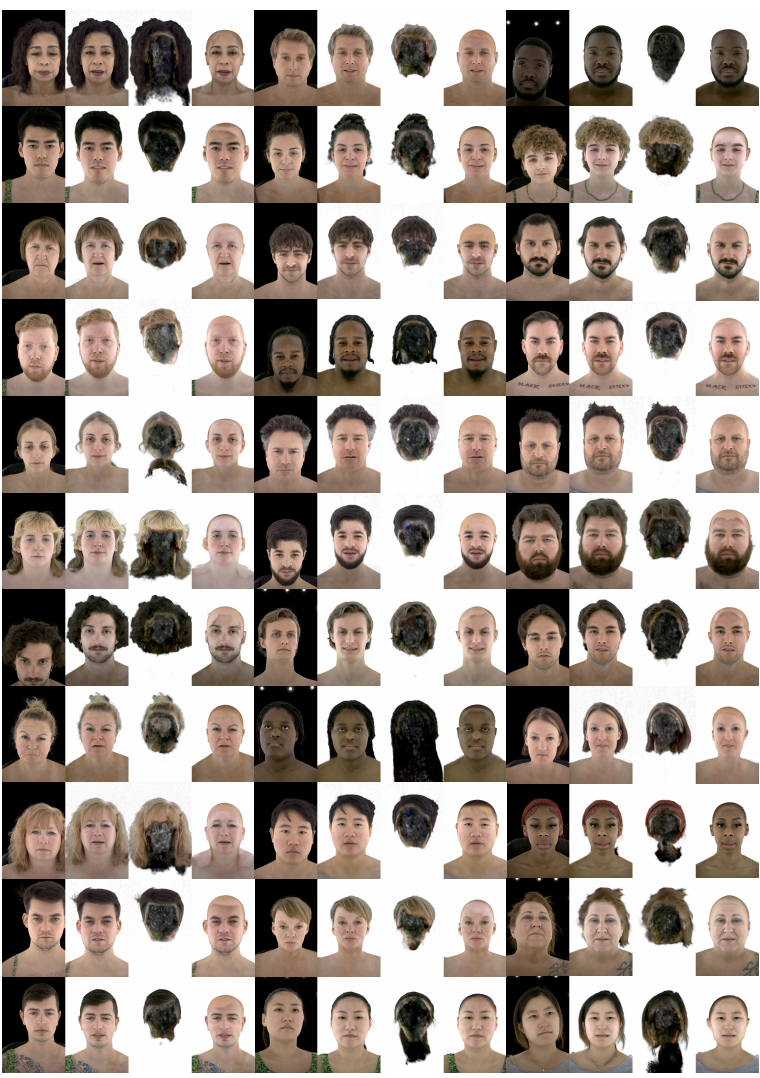}
   \caption{\textbf{Compositional 3D avatars of the training subjects.}
   }
   \label{fig:train_all}
\end{figure*}
\paragraph{Compositional 3D avatars.}
Our approach provides a unified 3D compositional representation of training subjects. \cref{fig:train_all} presents the results of our model trained with 64 subjects for compositional rendering, face-only rendering, and hair-only rendering, demonstrating effective separation of face and hair without compromising the quality of the combined 3D avatar. Notably, our model reconstructs a plausible facial appearance even in regions occluded by hair, which is crucial for seamless hairstyle transfer.

\begin{figure*}[t]
  \centering
   \includegraphics[width=0.7\linewidth]{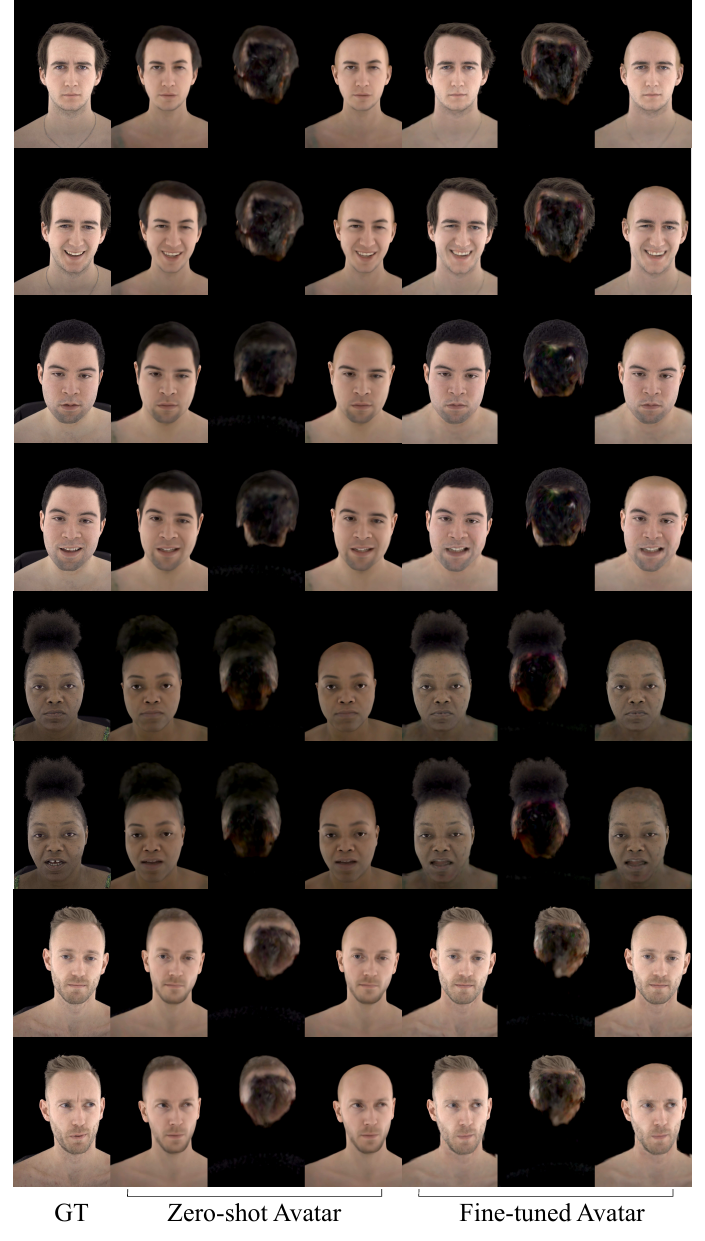}
   \caption{\textbf{Zero-shot and Fine-tuned Compositional Avatars.} Our model generates a plausible 3D avatar for a novel identity without fine-tuning (Zero-shot Avatar, middle column), reenacting the facial expression shown in the reference image (GT, left column). We visualize the compositional, hair-only, and face-only renderings for both the zero-shot and fine-tuned avatar results (Fine-tuned Avatar, right column). Fine-tuning improves visual fidelity and consistency while preserving the disentangled structure of face and hair.
   }
   \label{fig:zeroshot}
\end{figure*}

\paragraph{Zero-shot and fine-tuned 3D compositional avatars.}
Our model extends zero-shot inference to a compositional setting, generating 3D avatars with plausible face and hair representations by directly feeding the geometry and albedo maps of a novel identity into the identity-conditioned hypernetworks~\cite{cao2022authentic,li2024uravatar}, without requiring fine-tuning. Unlike autodecoder-based models~\cite{park2019deepsdf}, which require latent code inversion to obtain reasonable results for unseen identities~\cite{zheng2024headgap}, our hypernetwork-based design enables zero-shot inference in a simple feed-forward manner.
As shown in \cref{fig:zeroshot}, our zero-shot compositional avatar successfully reenacts expressions while reconstructing a full 3D appearance, even in regions originally occluded by hair, benefiting from the priors learned during pretraining. However, hair reconstruction in zero-shot results is less detailed compared to the face. This is because hair exists in a significantly higher-dimensional manifold with complex variations in shape and texture, making it more challenging to model. Fine-tuning~\cref{subsection:application} on a head rotation video with a neutral expression refines both facial and hair details, significantly enhancing visual fidelity.

\fi

\end{document}